\documentclass[lettersize,journal]{IEEEtran}
\usepackage{amsmath,amsfonts}
\usepackage{algorithmic}
\usepackage{algorithm}
\usepackage{array}
\usepackage[caption=false,font=normalsize,labelfont=sf,textfont=sf]{subfig}
\usepackage{textcomp}
\usepackage{stfloats}
\usepackage{url}
\usepackage{verbatim}
\usepackage{graphicx}
\usepackage[table,xcdraw]{xcolor}
\usepackage{multirow}
\usepackage[colorlinks=true, linkcolor=blue, citecolor=blue]{hyperref}
\usepackage{booktabs} 

\begin{document}

\title{LF-GNSS: Towards More Robust Satellite Positioning with a Hard Example Mining Enhanced Learning-Filtering Deep Fusion Framework}

\author
{
Jianan Lou, \IEEEmembership{Graduate Student Member, IEEE}, Rong Zhang
\thanks{Jianan Lou and Rong Zhang are with the Department of Precision Instrument and the State Key Laboratory of Precision Space-Time Information Sensing Technology, Tsinghua University, Beijing, 100084, China (e-mail: ljn22@mails.tsinghua.edu.cn; rongzh@mail.tsinghua.edu.cn).}
}

\markboth{Journal of \LaTeX\ Class Files,~Vol.~14, No.~8, August~2021}%
{Shell \MakeLowercase{\textit{et al.}}: A Sample Article Using IEEEtran.cls for IEEE Journals}

 \IEEEpubid{0000--0000/00\$00.00~\copyright~2021 IEEE}

\maketitle

\begin{abstract}
Global Navigation Satellite System (GNSS) is essential for autonomous driving systems, unmanned vehicles, and various location-based technologies, as it provides the precise geospatial information necessary for navigation and situational awareness. However, its performance is often degraded by Non-Line-Of-Sight (NLOS) and multipath effects, especially in urban environments. Recently, Artificial Intelligence (AI) has been driving innovation across numerous industries, introducing novel solutions to mitigate the challenges in satellite positioning. This paper presents a learning-filtering deep fusion framework for satellite positioning, termed LF-GNSS. The framework utilizes deep learning networks to intelligently analyze the signal characteristics of satellite observations, enabling the adaptive construction of observation noise covariance matrices and compensated innovation vectors for Kalman filter input. A dynamic hard example mining technique is incorporated to enhance model robustness by prioritizing challenging satellite signals during training. Additionally, we introduce a novel feature representation based on Dilution of Precision (DOP) contributions, which helps to more effectively characterize the signal quality of individual satellites and improve measurement weighting. LF-GNSS has been validated on both public and private datasets, demonstrating superior positioning accuracy compared to traditional methods and other learning-based solutions. To encourage further integration of AI and GNSS research, we will open-source the code at \underline{https://github.com/GarlanLou/LF-GNSS}, and release a collection of satellite positioning datasets for urban scenarios at \underline{https://github.com/GarlanLou/LF-GNSS-Dataset}.
\end{abstract}


\section{Introduction}
\label{sec:introduction}
\IEEEPARstart{G}{NSS} are essential for providing absolute positioning on Earth, underpinning advanced applications such as autonomous driving and smart cities \cite{zhang2024gnss}. However, satellite positioning systems face significant limitations. In particular, urban canyon environments are susceptible to interference from multipath effects and NLOS signals, which can severely degrade positioning accuracy \cite{groves2013portfolio}.

To address the aforementioned challenges, the integration of additional sensors has emerged as a promising solution. One widely adopted approach is GNSS/Inertial Navigation System (INS) integration \cite{li2024enhanced,shen2024novel,geng2024spoofing,geng2024covert}. These methods enhance the robustness of positioning systems by incorporating inertial information. However, when GNSS signals are lost for an extended period, the drift of the INS can cause the entire positioning system to fail. To address this issue, many approaches further integrate cameras into the framework \cite{song2024r2,liu2023ingvio,niu2022ic,cao2022gvins,li2022p,xia2024invariant}. These approaches leverage redundant information from additional sensors, significantly enhancing system performance. However, they also markedly increase the complexity and cost.

\begin{figure}[!t]
\centerline{\includegraphics[width=\linewidth,height=40mm]{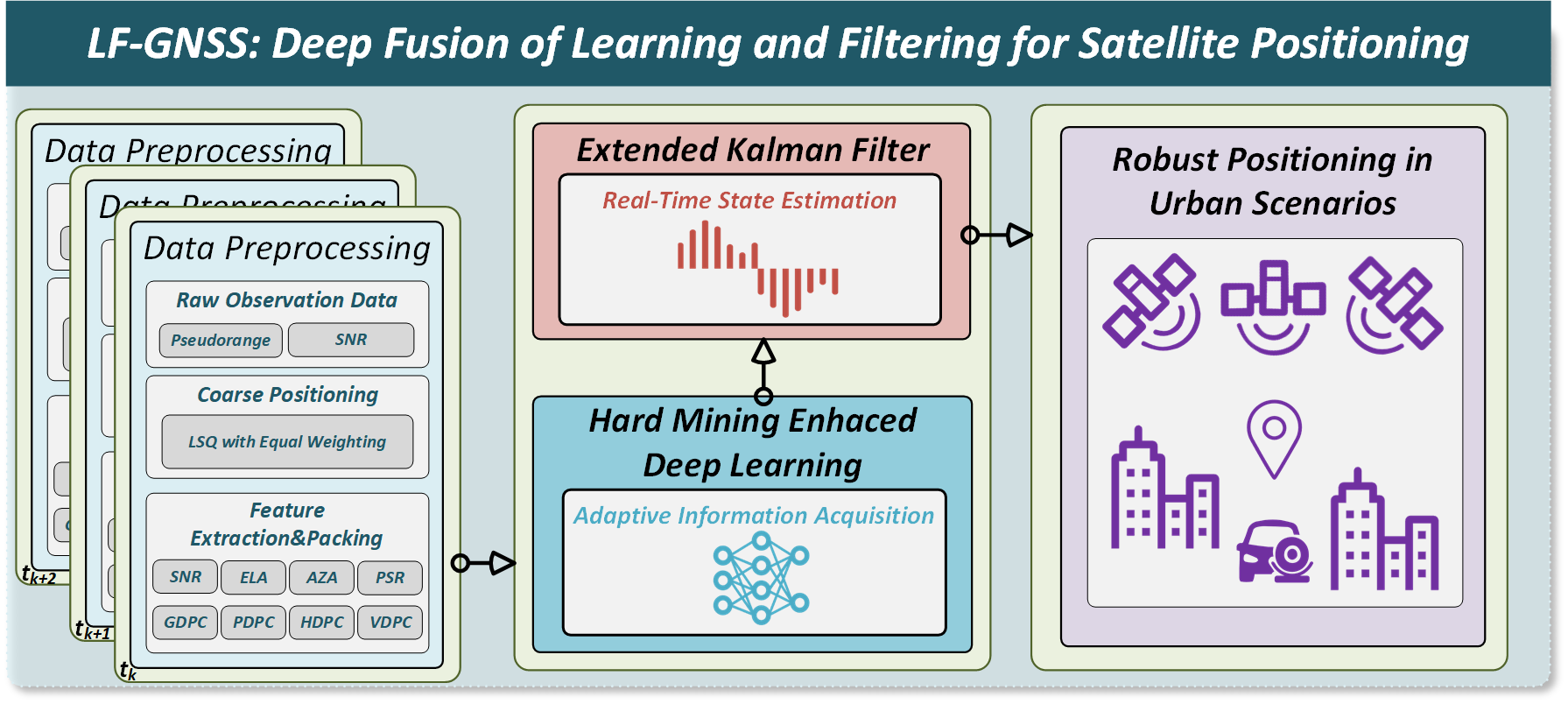}}
\caption{LF-GNSS: an open-sourced deep learning and Kalman filter integrated framework for satellite positioning.}
\label{Fig1_LF-GNSS}
\end{figure}

\IEEEpubidadjcol
Recent advancements in AI have significantly contributed to the development of various technologies, especially in autonomous driving \cite{chen2024end}. Machine Learning (ML) methods have demonstrated notable progress and outperformed traditional mathematical approaches in handling dynamic and complex problems, particularly under diverse environmental conditions and contexts. ML techniques excel at maintaining a balance between scalability and efficiency by accurately modeling intricate non-linear relationships among system variables \cite{mallik2023paving}. Enhancing GNSS positioning performance in urban environments through the integration of AI with GNSS, without relying on additional sensors, has become a promising research direction \cite{xu2024machine}. Siemuri et al. \cite{siemuri2021machine} have elaborated in detail on the application of AI in various aspects of GNSS.

Numerous scholars have combined traditional ML techniques with GNSS to boost its performance. Sun et al. \cite{sun2023resilient} applied random forest models to correct pseudorange errors in urban GNSS, resulting in enhanced horizontal accuracy. Zhang et al. \cite{zhang2024reliable} introduced a LightGBM-based method for GNSS NLOS error identification. However, it is worth noting that traditional ML methods may suffer from limitations compared to Deep Learning (DL) approaches. Specifically, the leaf-wise growth approach of LightGBM can lead to overfitting, particularly when dealing with small datasets or a limited number of features, which is one of the challenges that DL models are often better equipped to handle due to their stronger representation ability and adaptive feature selection.

The integration of DL and GNSS is driving interdisciplinary research to enhance satellite visibility prediction and error correction. Zhang et al. \cite{zhang2021prediction} used a DL model combining Fully Connected Neural Networks (FCNN) and Long Short-Term Memory (LSTM) networks to predict satellite visibility and pseudorange errors, demonstrating the ability of DL to extract environmental features. Zheng et al. \cite{zheng2024improving} proposed GTNN, a Graph Neural Network (GNN) with multi-head attention, to model satellite interactions and handle irregular GNSS measurements through graph-based aggregation. Xu et al. \cite{xu2024framework} transformed NLOS and multipath issues into graph-based tasks using the Single Differenced Residual (SDRes) map, addressing positioning, signal prediction, satellite weighting, and error computation, further integrating DL methods to mitigate GNSS errors.

Enhancing the reliability and smoothness of state estimation, i.e., developing robust estimators, is critical for navigation and positioning \cite{Zhang2022CTFGO}. While the aforementioned DL methods have achieved certain success in signal recognition and error correction, they have yet to establish a tight coupling between DL method and estimators.

Recent studies have begun to address this gap by integrating DL with traditional estimators to improve the robustness and accuracy of GNSS Positioning. Hu et al. \cite{hu2024pyrtklib} developed the open-source TDL-GNSS framework, achieving the coupling between fully connected layers and least squares estimation. Mohanty et al. \cite{mohanty2023learning} developed a hybrid GNSS positioning approach by integrating a Graph Convolutional Neural Network (GCNN) with a Kalman Filter (KF), later coupling a Graph Neural Network (GNN) with a Bayesian Kalman Filter (BKF) for refined smartphone-based corrections \cite{mohanty2024tightly}. Ding et al. \cite{ding2022learning} proposed LEAR-EKF, enhancing GNSS navigation by inferring Huber loss hyperparameters from satellite data, improving robustness in dynamic environments.

The above studies indicate that filtering algorithms, such as the Extended Kalman Filter (EKF), are widely employed in satellite positioning. Compared to graph optimization algorithms, EKF offers lower computational overhead. However, the adaptive determination of process noise and measurement noise remains a significant challenge. Proper tuning of these noise parameters is critical for the performance of EKF, as inappropriate values can lead to poor estimation accuracy or even filter divergence \cite{revach2022kalmannet}. Extracting features from satellite signals and tightly coupling these features with estimators offers a compelling approach to enhancing positioning performance.

Recognizing both the potential of integrating AI with GNSS and the existing limitations in this field, we propose a learning-filtering fusion satellite positioning framework designed to improve positioning accuracy in complex urban environments. The main contributions of this paper are summarized as follows:

\textit{1) Fusion of Learning and Filtering:}  
We introduce an innovative framework that seamlessly integrates learning-based methodologies with advanced filtering techniques to revolutionize satellite positioning in urban environments. This synergy harnesses the complementary advantages of both paradigms, facilitating adaptive noise modeling, mitigating NLOS errors, and ensuring robust and high-precision state estimation under challenging conditions.

\textit{2) DOP-based Feature Representation:}
We propose a novel DOP-based feature representation that explicitly captures the geometric contribution of each satellite. By leveraging DOP as a key feature in our learning framework, we enhance the ability to characterize satellite signal quality and refine measurement weighting strategies. This advancement enables more accurate and adaptive fusion of multi-GNSS observations, particularly in challenging urban environments where satellite visibility and signal reliability fluctuate dynamically.

\textit{3) Dynamic Hard Example Mining:}  
We propose a dynamic hard example mining algorithm, recognizing the varying difficulties of satellite positioning across different environments. This algorithm further improves the robustness and accuracy of the framework by selectively focusing on difficult samples during training, ensuring better adaptation to challenging urban scenarios.

\textit{4) Validation and Performance Improvement:}  
We validated our framework on a wide range of public and private datasets, showing that positioning accuracy is improved compared to traditional frameworks such as CSSRLIB, goGPS and RTKLIB, as well as open-source DL-based frameworks like TDL-GNSS. The experiments also demonstrate that the pre-trained model, when compared to higher-precision positioning methods such as Precise Point Positioning (PPP), exhibits better stability in urban scenarios. Ablation studies further confirm the effectiveness of our innovations, highlighting the significant contributions of each component to the overall performance improvement.

\textit{5) Open Source Contribution and Dataset Release:}  
We will open-source the code and also release a dataset of satellite data in urban scenarios. We believe this will contribute positively to the research on the integration of AI and satellite positioning algorithms.

\begin{figure*}[!t]
\centerline{\includegraphics[width=\textwidth]{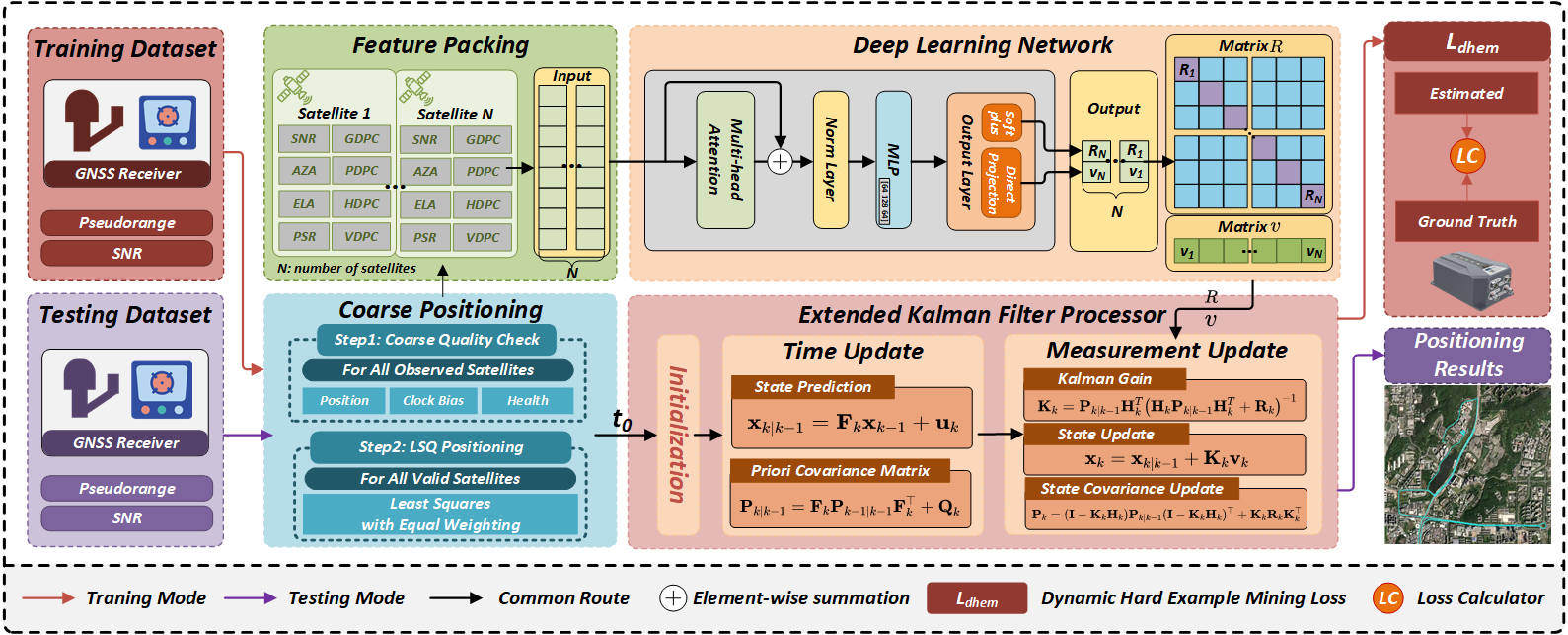}}
\caption{System overview of the proposed LF-GNSS framework. It consists of four main modules: Coarse Positioning, Feature Packing, Deep Learning Network and Extended Kalman Filter Processor.}
\label{Fig2_SystemOverview}
\end{figure*}

\section{System Overview}
The overall process of the proposed LF-GNSS framework is illustrated in Fig. \ref{Fig2_SystemOverview}. This framework consists of four main modules:

\textit{1) Coarse Positioning:}  
This module involves two steps. The first step is coarse quality control, which filters out gross errors to improve stability. The second step applies equal-weighted least squares, serving the dual purpose of initializing the first epoch and extracting features from the positioning results.

\textit{2) Feature Packing:}  
This module normalizes and processes all features, structuring them into a unified format. As the number of satellites observed varies at each epoch, feature information from different epochs is expanded and transformed to maintain a consistent dimension, ensuring uniform input to the subsequent DL model.

\textit{3) DL Network:}  
Feature values representing satellite observation quality are fed into the DL network. A multi-head attention mechanism is employed to capture intrinsic relationships between satellite features, allowing the model to dynamically weigh and differentiate between observations. The network outputs the measurement noise covariance matrix \( \mathbf{R} \), characterizing measurement noise levels, and the vector \( \mathbf{v}^c \), which represents the compensated innovation and directly encapsulates observation information at the current epoch.

\textit{4) EKF Processor:}  
The EKF receives the matrices \( \mathbf{R} \) and \( \mathbf{v}^c \) from the network to participate in the filtering process, yielding positioning results for the current epoch.

Our framework operates in two distinct modes, addressing different stages of deployment and application. These modes allow the framework to learn from data and continuously improve, while also enabling real-time GNSS positioning once the model is trained and optimized.

\textit{1) Training Mode:}  
In training mode, the positioning results are compared with ground truth to compute the error. A dynamic hard example mining strategy is employed to construct the loss function, which is then used to update and optimize the network. This iterative process allows our framework to adapt to diverse environments.

\textit{2) Testing Mode:}  
In testing mode, the trained model intelligently captures satellite signal features and adaptively adjusts the Kalman filter, enabling real-time GNSS positioning with improved accuracy and stability.

\renewcommand{\arraystretch}{1.25}

\section{METHODOLOGY}
\label{sec:methodology}
\subsection{Coarse Positioning}
The aim of the coarse positioning module is to obtain the approximate position of the GNSS antenna while extracting certain signal features. Given the instability of carrier phase measurements in complex urban scenarios and the cumbersome ambiguity resolution process, we exclusively employ undifferenced pseudorange observations in our framework, which is simple, efficient, and easy to implement.

Numerous studies have demonstrated that the accuracy of pseudorange is sufficient to distinguish between roads in urban environments \cite{niu2024mgins}. The pseudorange measurement \( P_r^s \) between a satellite (denoted as \( s \)) and a receiver (denoted as \( r \)) represents the approximate distance information between the satellite and the receiver. The pseudorange measurement equation is given as:
\begin{equation}
P_r^s = c \cdot [t_r(t) - t^s(t - \tau^s)] + e_r^s(t)
\end{equation}
where:
\begin{itemize}
    \item \( \Delta t_r \): Receiver clock offset.  
    \item \( \Delta t^s \): Satellite clock offset.  
    \item \( \tau^s \): Signal transmission time.  
    \item \( t_r(t) \): Time of reception at the receiver.  
    \item \( t^s(t - \tau^s) \): Time of transmission at the satellite.  
    \item \( c \): Speed of light.  
    \item \( e_r^s(t) \): Pseudorange measurement noise.  
\end{itemize}

The signal transmission time \( \tau^s \) can be expressed as:
\begin{align}
\tau^s &= \tau_r^s + \text{dcb}_r + \text{dcb}^s \notag \\
       &= \frac{1}{c} \left( \rho_r^s + I_r^s + T_r^s + dm_r^s \right) + \text{dcb}^s
\end{align}
where:
\begin{itemize}
    \item \( \tau_r^s \): Actual signal transmission time.
    \item \( \text{dcb}_r \): Receiver hardware delay.
    \item \( \text{dcb}^s \): Satellite hardware delay.
    \item \( \rho_r^s \): True geometric range between the satellite and the receiver.
    \item \( I_r^s \): Ionospheric delay.
    \item \( T_r^s \): Tropospheric delay.
    \item \( dm_r^s \): Multipath error.
\end{itemize}

Combining the equations above, the comprehensive pseudorange measurement equation is derived as follows:
\begin{align}
P_r^s &= \rho_r^s + c \cdot \Delta t_r - c \cdot \Delta t^s \notag \\
      &\quad + c \cdot \text{dcb}_r + c \cdot \text{dcb}^s + I_r^s + T_r^s + dm_r^s + e_r^s(t)
\label{eq:pseudorange}
\end{align}

Based on the pseudorange observation Equation \ref{eq:pseudorange}, after applying the Saastamoinen model \cite{bevis1994gps} to correct for tropospheric delay and the Klobuchar model \cite{klobuchar1987ionospheric} to mitigate ionospheric delay, eliminating satellite clock errors using ephemeris data, and accounting for hardware delays via the Time Group Delay (TGD) correction and other minor errors, the equation simplifies to:
\begin{equation}
P_r^s = \rho_r^s + c \cdot \Delta t_r
\end{equation}

Assume the satellite position is \( (x_s, y_s, z_s) \) and the receiver position is \( (x_r, y_r, z_r) \). The pseudorange can be expressed as:
\begin{multline}
P_{r}^s = \sqrt{(x_s - x_r)^2 + (y_s - y_r)^2 + (z_s - z_r)^2}+ \Delta t_r \\
 = f(x_r, y_r, z_r, \Delta t_r)
\end{multline}

When there are \( N \) satellites at a given observation epoch, the pseudorange observation equation group is:
\begin{equation}
\begin{bmatrix}
P_{1r}^s \\
P_{2r}^s \\
P_{3r}^s \\
\vdots \\
P_{Nr}^s
\end{bmatrix}
=
\begin{bmatrix}
f_1(x_r, y_r, z_r, \Delta t_r) \\
f_2(x_r, y_r, z_r, \Delta t_r) \\
f_3(x_r, y_r, z_r, \Delta t_r) \\
\vdots \\
f_N(x_r, y_r, z_r, \Delta t_r)
\end{bmatrix}
\end{equation}

This nonlinear equation can be solved using the least squares method. The pseudorange observation equation group can be written in vector form as:  
\begin{equation}
\mathbf{z} = f(\mathbf{x})
\label{z}
\end{equation}
where \( \mathbf{x} = \left( x_r, y_r, z_r, \Delta t_r \right)^\top \) represents the receiver's coordinates and clock offset, and \( \mathbf{z} \) denotes the pseudorange observation vector.

According to the nonlinear least squares optimization method, the solution for \( \mathbf{x} \) is obtained by minimizing the residual:
\begin{equation}
\mathbf{l} = f(\mathbf{x}) - \mathbf{z}, \quad \text{minimize } \|\mathbf{l}\|
\end{equation}
where \( \mathbf{z} \) represents the vector of observed pseudorange measurements, and \( f(\mathbf{x}) \) is the vector of modeled pseudoranges based on the estimated receiver state \( \mathbf{x} \).

The equation presented describes the iterative least squares (ILS) method used in GNSS positioning, particularly for solving nonlinear least squares problems by gradually approximating the optimal solution through iterative updates.
\begin{equation}
\begin{cases}
\mathbf{x}_{k+1} = \mathbf{x}_k + \Delta \mathbf{x} \\
\Delta \mathbf{x} = \left[\mathbf{H}^T \mathbf{W} \mathbf{H}\right]^{-1} \cdot \mathbf{H}^T \cdot \mathbf{l}
\end{cases}
\end{equation}

\noindent where:
\begin{itemize}
    \item \( \mathbf{x}_k \): Represents the state vector at the \( k \)-th iteration.
    \item \( \mathbf{x}_{k+1} \): Represents the updated state vector at iteration \( k+1 \), obtained by adding the correction term \( \Delta \mathbf{x} \) to the current state \( \mathbf{x}_k \).
    \item \( \Delta \mathbf{x} \): Represents the state increment, calculated using the weighted least squares solution:
    \begin{itemize}
        \item \( \mathbf{H} \): The Jacobian matrix, representing the partial derivatives of the system with respect to the state vector.
        \item \( \mathbf{W} \): The weighting matrix, typically the inverse of the observation noise covariance matrix.
        \item \( \mathbf{l} \): The pseudorange residual vector, representing the difference between observed and modeled pseudoranges.
    \end{itemize}
\end{itemize}

The Jacobian matrix \( \mathbf{H} \) for the system with \( N \) satellites can be expressed as:
\begin{equation}
\mathbf{H}(\mathbf{x}) =
\begin{bmatrix}
\frac{\partial f_1}{\partial x} & \frac{\partial f_1}{\partial y} & \frac{\partial f_1}{\partial z} & \frac{\partial f_1}{\partial \Delta t_r} \\
\frac{\partial f_2}{\partial x} & \frac{\partial f_2}{\partial y} & \frac{\partial f_2}{\partial z} & \frac{\partial f_2}{\partial \Delta t_r} \\
\vdots & \vdots & \vdots & \vdots \\
\frac{\partial f_N}{\partial x} & \frac{\partial f_N}{\partial y} & \frac{\partial f_N}{\partial z} & \frac{\partial f_N}{\partial \Delta t_r}
\end{bmatrix}
\end{equation}

For simplicity, let:
\begin{equation}
D = \sqrt{(x^{s} - x_r)^2 + (y^{s} - y_r)^2 + (z^{s} - z_r)^2}.
\label{eq:D}
\end{equation}
Then the Jacobian matrix can be expressed as:
\begin{equation}
\mathbf{H}(\mathbf{x}) =
\begin{bmatrix}
-\frac{x_1^s - x_r}{D_1} & -\frac{y_1^s - y_r}{D_1} & -\frac{z_1^s - z_r}{D_1} & 1 \\
-\frac{x_2^s - x_r}{D_2} & -\frac{y_2^s - y_r}{D_2} & -\frac{z_2^s - z_r}{D_2} & 1 \\
\vdots & \vdots & \vdots & \vdots \\
-\frac{x_N^s - x_r}{D_N} & -\frac{y_N^s - y_r}{D_N} & -\frac{z_N^s - z_r}{D_N} & 1
\end{bmatrix}
\label{eq:H Matrix}
\end{equation}


To fully reflect the observation conditions of each satellite without introducing any prior assumptions, we use the identity matrix as the weighting matrix. The weighting matrix \( \mathbf{W} \) for \( N \) satellites is expressed as \( \mathbf{W} = \mathbf{I}_N \), where \( \mathbf{I}_N \) denotes an \( N \times N \) identity matrix.

LF-GNSS supports multiple constellations, so it is necessary to estimate the Inter-System Biases (ISB) between different constellations. The state vector is formulated as:
\begin{equation}
\label{state vector 1}
\mathbf{x} = \left( x_r, y_r, z_r, \Delta t_r, \mathbf{ISB} \right)^\top
\end{equation}

The Jacobian matrix \( \mathbf{H} \), accommodating multiple clock biases, can be readily derived through the transformation of Equation \ref{eq:H Matrix}.

\begin{equation}
\label{H}
\mathbf{H(x)} =
\begin{bmatrix}
-\frac{x_1^s - x_r}{D_1} & -\frac{y_1^s - y_r}{D_1} & -\frac{z_1^s - z_r}{D_1} & 1 & \alpha_1^C & \alpha_1^E & \alpha_1^R \\
-\frac{x_2^s - x_r}{D_2} & -\frac{y_2^s - y_r}{D_2} & -\frac{z_2^s - z_r}{D_2} & 1 & \alpha_2^C & \alpha_2^E & \alpha_2^R \\
\vdots & \vdots & \vdots & \vdots & \vdots & \vdots & \vdots \\
-\frac{x_N^s - x_r}{D_N} & -\frac{y_N^s - y_r}{D_N} & -\frac{z_N^s - z_r}{D_N} & 1 & \alpha_N^C & \alpha_N^E & \alpha_N^R \\
\end{bmatrix}
\end{equation}

where:
\begin{itemize}
    \item If a satellite belongs to GPS, then \( \alpha^C, \alpha^E, \alpha^R \) are all set to zero.
    \item If a satellite belongs to BeiDou, then \( \alpha^C = 1 \), while \( \alpha^E \) and \( \alpha^R \) remain zero.
    \item If a satellite belongs to Galileo, then \( \alpha^E = 1 \), while \( \alpha^C \) and \( \alpha^R \) are zero.
    \item If a satellite belongs to GLONASS, then \( \alpha^R = 1 \), while \( \alpha^C \) and \( \alpha^E \) are zero.
\end{itemize}

\begin{figure*}[!t]
\centerline{\includegraphics[width=180mm,height=100mm]{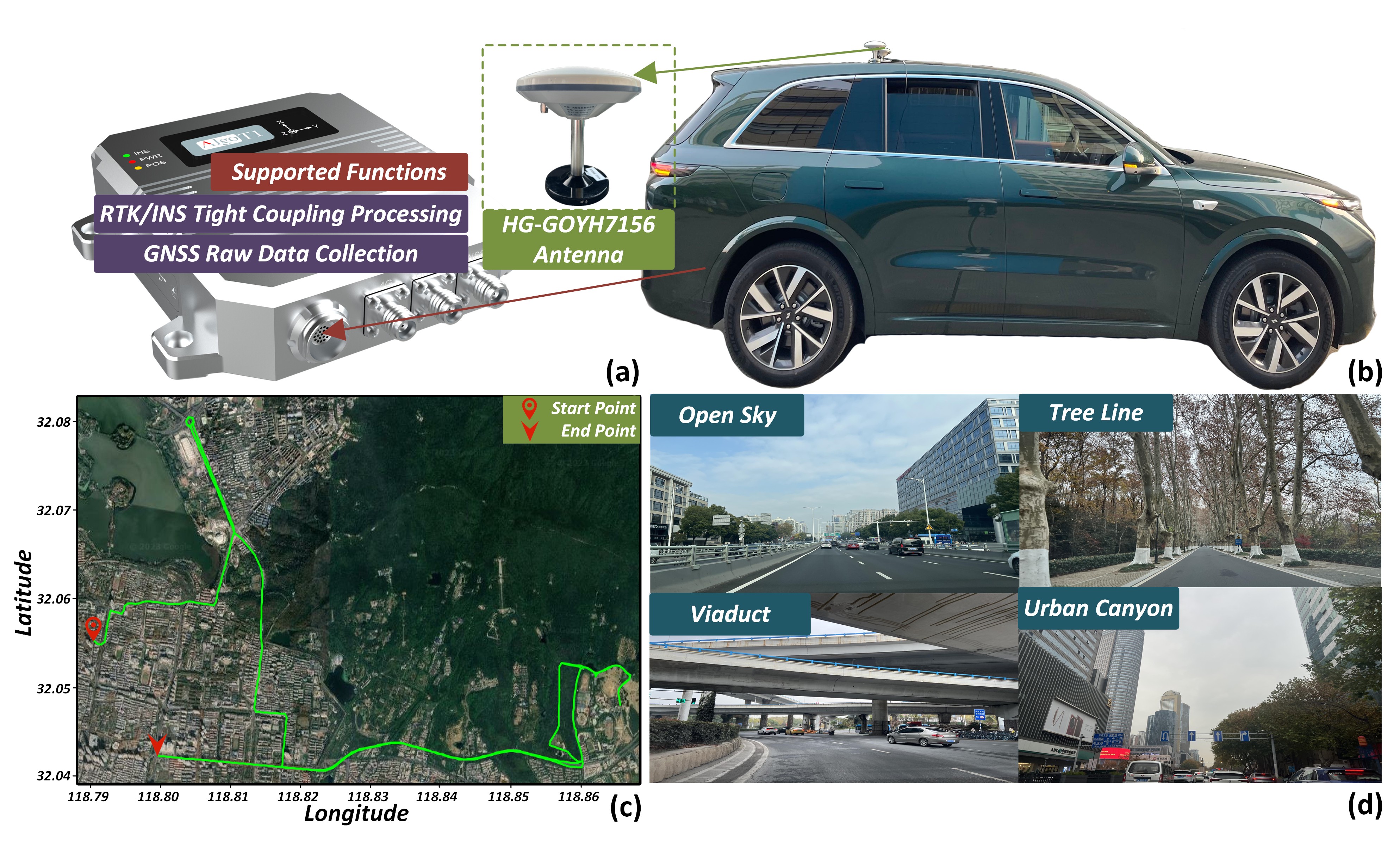}}
\caption{Private experiment details: platform, test route and scenarios.}
\label{Fig6_PrivateExperimentDetails}
\end{figure*}

\subsection{Feature Packing}
Based on previous work \cite{zhang2024reliable}, \cite{zhang2024ins}, \cite{zhang2023learning}, we selected four elements to represent the observation quality of each satellite at the current epoch. In addition, we introduce a fifth feature, which is proposed in this study to further enhance the representation. The implementation of the coarse positioning module facilitates the efficient extraction of all these features.

\subsubsection{Signal-to-Noise Ratio (SNR)}  
The Signal-to-Noise Ratio (SNR) reflects the strength and quality of satellite signals and serves as a key indicator of observation reliability. It can be directly obtained from raw observations, with higher values indicating stronger and more stable GNSS signals. In urban environments, SNR is an effective metric for detecting signal degradation caused by NLOS receptions and multipath effects.

\subsubsection{Elevation Angle (ELA)}  
The satellite elevation angle (\(\text{ELA}\)) is a critical measure of satellite visibility and signal quality. Satellites with higher elevation angles are less susceptible to obstructions and multipath effects, making them more reliable for GNSS positioning. In our framework, the coarse positioning module facilitates the calculation of the satellite elevation angle by utilizing the estimated receiver position. The elevation angle \(\text{ELA}_n\) for the \(n\)-th satellite is calculated as:
\begin{equation}
\text{ELA}_n = \arcsin\left(\frac{U^{s}_{n}}{D_n}\right),
\end{equation}
where \( U^{s}_{n} \) represents the satellite's height relative to the receiver in the East-North-Up (ENU) frame, derived from the coarse position estimate. The term \( D_n \) indicates the geometric range between the receiver and the \(n\)-th satellite, computed as Equation \ref{eq:D}.
 
\subsubsection{Azimuth Angle (AZA)}  
The satellite azimuth angle (\(\text{AZA}\)) represents the horizontal geometric relationship between the satellite and the receiver. Azimuth angles can help infer satellite visibility, especially when multiple satellites with similar azimuth angles but different elevation angles are observed. This feature enhances the geometric context in learning-based methods. The azimuth angle for the \( n \)-th satellite is calculated as:  
\begin{equation}
\text{AZA}_n = \arctan\left(\frac{x^{s}_{n}}{y^{s}_{n}}\right),
\end{equation}
where \( x^{s}_{n} \) and \( y^{s}_{n} \) represent the position of the \( n \)-th satellite relative to the receiver in the ENU frame.

\subsubsection{Pseudorange Residual (PSR)}  
In our framework, the vehicle position is initially estimated using an ILS method in the coarse positioning module. This solution is then used to compute pseudorange residuals, denoted as \( \text{PSR}_n \), which can directly reflect GNSS observation quality and the magnitude of pseudorange errors. The least-squares pseudorange residual is calculated as:
\begin{equation}
\text{PSR}_n = P_{nr}^{s} - D_n,
\end{equation}
where \( P_{nr}^{s} \) represents the pseudorange measurement between the receiver and the \( n \)-th satellite. The geometric distance \( D_n \) can also be computed as Equation \ref{eq:D}. 

\subsubsection{Dilution of Precision Contribution (DPC)}  
DOP is a key indicator reflecting the geometric distribution of satellites and its influence on GNSS positioning accuracy. A higher DOP value indicates poor satellite geometry, leading to greater positioning errors. The DOP is commonly decomposed into four main components:

\begin{itemize}
    \item Geometric DOP (GDOP): Overall positioning accuracy, accounting for spatial dimensions and clock bias.  
    \item Position DOP (PDOP): Contribution of horizontal and vertical components to positioning precision.  
    \item Horizontal DOP (HDOP): Horizontal positioning error along the East and North axes.  
    \item Vertical DOP (VDOP): Vertical positioning accuracy along the Up axis.  
\end{itemize}

In our framework, DOP is computed based on AZA and ELA of observed satellites. The design matrix \( \mathbf{H} \) describes the geometric relationship between the receiver and each satellite. For \( N \) satellites, the design matrix \( \mathbf{H} \) is formulated as:
\begin{equation}
\begin{bmatrix}
\text{c}(\text{ELA}_1)\text{s}(\text{AZA}_1) & \text{c}(\text{ELA}_1)\text{c}(\text{AZA}_1) & \text{s}(\text{ELA}_1) & 1 \\
\text{c}(\text{ELA}_2)\text{s}(\text{AZA}_2) & \text{c}(\text{ELA}_2)\text{c}(\text{AZA}_2) & \text{s}(\text{ELA}_2) & 1 \\
\vdots & \vdots & \vdots & \vdots \\
\text{c}(\text{ELA}_N)\text{s}(\text{AZA}_N) & \text{c}(\text{ELA}_N)\text{c}(\text{AZA}_N) & \text{s}(\text{ELA}_N) & 1
\end{bmatrix}
\end{equation}
where \(\text{s}(\cdot)\) and \(\text{c}(\cdot)\) represent \(\sin(\cdot)\) and \(\cos(\cdot)\), respectively.

The inverse of the normal matrix is calculated as:
\begin{equation}
\mathbf{Q} = (\mathbf{H}^T \mathbf{H})^{-1}
\end{equation}

The diagonal elements of \( \mathbf{Q} \) yield the DOP components:
\begin{equation}
\text{HDOP} = \sqrt{\mathbf{Q}_{11} + \mathbf{Q}_{22}}
\end{equation}
\vspace{-0.5cm} 
\begin{equation}
\text{VDOP} = \sqrt{\mathbf{Q}_{33}}
\end{equation}
\vspace{-0.5cm} 
\begin{equation}
\text{PDOP} = \sqrt{\mathbf{Q}_{11} + \mathbf{Q}_{22} + \mathbf{Q}_{33}}
\end{equation}
\vspace{-0.5cm} 
\begin{equation}
\text{GDOP} = \sqrt{\text{PDOP}^2 + \mathbf{Q}_{44}}
\end{equation}

We introduce the concept of DOP contribution as a novel feature to assess the impact of each satellite on overall DOP. This is achieved by iteratively removing satellites from the design matrix and recalculating DOP values. The difference between full and reduced DOP quantifies the influence of the excluded satellite. This approach identifies satellites that degrade positioning accuracy, allowing for selective exclusion to enhance GNSS performance, particularly in urban environments prone to signal obstructions and multipath interference.
\begin{equation}
\text{DPC}_n = \Delta \text{DOP} = \text{DOP}_{\text{full}} - \text{DOP}_{\text{partial}}
\end{equation}
where \( \text{DOP}_{\text{full}} \) represents DOP with all satellites, and \( \text{DOP}_{\text{partial}} \) is calculated by excluding the \( n \)-th satellite.

\subsection{Deep Learning Network Design}  
The core of the deep learning component in our framework is a sophisticated fully connected neural network augmented with multi-head self-attention, designed to estimate the measurement noise covariance matrix \( \mathbf{R} \) and the innovation compensation vector \( \mathbf{v}^{\mathbf{c}} \) for GNSS observations. This advanced architecture effectively captures complex interdependencies among satellite signals, while ensuring computational efficiency, thus enabling seamless integration into real-time positioning systems with high precision and adaptability.

\begin{table*}[]
\centering
\caption{Public Dataset Experiment Results [RMSE, m]}
\label{Public Dataset Experiments Results}
\renewcommand{\arraystretch}{1.7} 
\setlength{\tabcolsep}{0.8pt}
\begin{tabular}{
>{\columncolor[HTML]{FFFFFF}}c 
>{\columncolor[HTML]{FFFFFF}}c 
>{\columncolor[HTML]{FFFFFF}}c 
>{\columncolor[HTML]{FFFFFF}}c 
>{\columncolor[HTML]{FFFFFF}}c 
>{\columncolor[HTML]{FFFFFF}}c 
>{\columncolor[HTML]{FFFFFF}}c 
>{\columncolor[HTML]{FFFFFF}}c 
>{\columncolor[HTML]{FFFFFF}}c 
>{\columncolor[HTML]{FFFFFF}}c 
>{\columncolor[HTML]{FFFFFF}}c 
>{\columncolor[HTML]{FFFFFF}}c 
>{\columncolor[HTML]{FFFFFF}}c 
>{\columncolor[HTML]{FFFFFF}}c 
>{\columncolor[HTML]{FFFFFF}}c 
>{\columncolor[HTML]{FFFFFF}}c 
>{\columncolor[HTML]{FFFFFF}}c 
>{\columncolor[HTML]{FFFFFF}}c 
>{\columncolor[HTML]{FFFFFF}}c 
>{\columncolor[HTML]{FFFFFF}}c 
>{\columncolor[HTML]{FFFFFF}}c 
>{\columncolor[HTML]{FFFFFF}}c 
>{\columncolor[HTML]{F7E0DF}}c 
>{\columncolor[HTML]{F7E0DF}}c 
>{\columncolor[HTML]{F7E0DF}}c 
>{\columncolor[HTML]{F7E0DF}}c 
>{\columncolor[HTML]{F7E0DF}}c }
\toprule[1.3pt]
\multicolumn{2}{c}{\cellcolor[HTML]{FFFFFF}{\color[HTML]{000000} \textbf{Methods}}} &
  \multicolumn{5}{c}{\cellcolor[HTML]{FFFFFF}{\color[HTML]{000000} \textbf{CSSRLIB}}} &
  \multicolumn{5}{c}{\cellcolor[HTML]{FFFFFF}{\color[HTML]{000000} \textbf{goGPS}}} &
  \multicolumn{5}{c}{\cellcolor[HTML]{FFFFFF}{\color[HTML]{000000} \textbf{RTKLIB}}} &
  \multicolumn{5}{c}{\cellcolor[HTML]{FFFFFF}{\color[HTML]{000000} \textbf{TDL-GNSS}}} &
  \multicolumn{5}{c}{\cellcolor[HTML]{F7E0DF}{\color[HTML]{000000} \textbf{LF-GNSS (Ours)}}} \\ \midrule[1.1pt]
\multicolumn{2}{c!{\vrule width 1.1pt}}{\cellcolor[HTML]{FFFFFF}{\color[HTML]{000000} Metric}} &
  {\color[HTML]{000000} E} &
  {\color[HTML]{000000} N} &
  {\color[HTML]{000000} U} &
  {\color[HTML]{000000} 2D} &
  \multicolumn{1}{c!{\vrule width 1.1pt}}{\cellcolor[HTML]{FFFFFF}{\color[HTML]{000000} 3D}} &
  {\color[HTML]{000000} E} &
  {\color[HTML]{000000} N} &
  {\color[HTML]{000000} U} &
  {\color[HTML]{000000} 2D} &
  \multicolumn{1}{c!{\vrule width 1.1pt}}{\cellcolor[HTML]{FFFFFF}{\color[HTML]{000000} 3D}} &
  {\color[HTML]{000000} E} &
  {\color[HTML]{000000} N} &
  {\color[HTML]{000000} U} &
  {\color[HTML]{000000} 2D} &
  \multicolumn{1}{c!{\vrule width 1.1pt}}{\cellcolor[HTML]{FFFFFF}{\color[HTML]{000000} 3D}} &
  {\color[HTML]{000000} E} &
  {\color[HTML]{000000} N} &
  {\color[HTML]{000000} U} &
  {\color[HTML]{000000} 2D} &
  \multicolumn{1}{c!{\vrule width 1.1pt}}{\cellcolor[HTML]{FFFFFF}{\color[HTML]{000000} 3D}} &
  {\color[HTML]{000000} E} &
  {\color[HTML]{000000} N} &
  {\color[HTML]{000000} U} &
  {\color[HTML]{000000} 2D} &
  {\color[HTML]{000000} 3D} \\ \midrule[1.1pt]
\multicolumn{1}{c!{\vrule width 1.1pt}}{\cellcolor[HTML]{FFFFFF}{\color[HTML]{000000} }} &
  \multicolumn{1}{c!{\vrule width 1.1pt}}{\cellcolor[HTML]{FFFFFF}{\color[HTML]{000000} A}} &
  {\color[HTML]{000000} 1.51} &
  {\color[HTML]{000000} 1.68} &
  {\color[HTML]{000000} 9.08} &
  {\color[HTML]{000000} 2.26} &
  \multicolumn{1}{c!{\vrule width 1.1pt}}{\cellcolor[HTML]{FFFFFF}{\color[HTML]{000000} 9.35}} &
  {\color[HTML]{000000} 1.24} &
  {\color[HTML]{000000} 1.70} &
  {\color[HTML]{000000} 18.65} &
  {\color[HTML]{000000} 2.11} &
  \multicolumn{1}{c!{\vrule width 1.1pt}}{\cellcolor[HTML]{FFFFFF}{\color[HTML]{000000} 18.77}} &
  {\color[HTML]{000000} 2.18} &
  {\color[HTML]{000000} 1.58} &
  {\color[HTML]{000000} 10.21} &
  {\color[HTML]{000000} 2.69} &
  \multicolumn{1}{c!{\vrule width 1.1pt}}{\cellcolor[HTML]{FFFFFF}{\color[HTML]{000000} 10.56}} &
  {\color[HTML]{000000} 1.60} &
  {\color[HTML]{000000} 1.41} &
  {\color[HTML]{000000} 5.19} &
  {\color[HTML]{000000} 2.13} &
  \multicolumn{1}{c!{\vrule width 1.1pt}}{\cellcolor[HTML]{FFFFFF}{\color[HTML]{000000} 5.60}} &
  {\color[HTML]{000000} \textbf{1.11}} &
  {\color[HTML]{000000} \textbf{1.00}} &
  {\color[HTML]{000000} \textbf{3.99}} &
  {\color[HTML]{000000} \textbf{1.50}} &
  {\color[HTML]{000000} \textbf{4.26}} \\
\multicolumn{1}{c!{\vrule width 1.1pt}}{\cellcolor[HTML]{FFFFFF}{\color[HTML]{000000} }} &
  \multicolumn{1}{c!{\vrule width 1.1pt}}{\cellcolor[HTML]{FFFFFF}{\color[HTML]{000000} B}} &
  {\color[HTML]{000000} 3.15} &
  {\color[HTML]{000000} 2.51} &
  {\color[HTML]{000000} 8.82} &
  {\color[HTML]{000000} 4.03} &
  \multicolumn{1}{c!{\vrule width 1.1pt}}{\cellcolor[HTML]{FFFFFF}{\color[HTML]{000000} 9.70}} &
  {\color[HTML]{000000} 2.65} &
  {\color[HTML]{000000} 1.81} &
  {\color[HTML]{000000} 15.81} &
  {\color[HTML]{000000} 3.21} &
  \multicolumn{1}{c!{\vrule width 1.1pt}}{\cellcolor[HTML]{FFFFFF}{\color[HTML]{000000} 16.13}} &
  {\color[HTML]{000000} 3.60} &
  {\color[HTML]{000000} 3.57} &
  {\color[HTML]{000000} 11.76} &
  {\color[HTML]{000000} 5.06} &
  \multicolumn{1}{c!{\vrule width 1.1pt}}{\cellcolor[HTML]{FFFFFF}{\color[HTML]{000000} 12.80}} &
  {\color[HTML]{000000} 1.80} &
  {\color[HTML]{000000} 1.26} &
  {\color[HTML]{000000} 4.52} &
  {\color[HTML]{000000} 2.20} &
  \multicolumn{1}{c!{\vrule width 1.1pt}}{\cellcolor[HTML]{FFFFFF}{\color[HTML]{000000} 5.02}} &
  {\color[HTML]{000000} \textbf{1.43}} &
  {\color[HTML]{000000} \textbf{1.04}} &
  {\color[HTML]{000000} \textbf{3.62}} &
  {\color[HTML]{000000} \textbf{1.77}} &
  {\color[HTML]{000000} \textbf{4.02}} \\
\multicolumn{1}{c!{\vrule width 1.1pt}}{\multirow{-3}{*}{\cellcolor[HTML]{FFFFFF}{\color[HTML]{000000} IPNL}}} &
  \multicolumn{1}{c!{\vrule width 1.1pt}}{\cellcolor[HTML]{FFFFFF}{\color[HTML]{000000} C}} &
  {\color[HTML]{000000} 13.82} &
  {\color[HTML]{000000} 16.21} &
  {\color[HTML]{000000} 59.92} &
  {\color[HTML]{000000} 21.30} &
  \multicolumn{1}{c!{\vrule width 1.1pt}}{\cellcolor[HTML]{FFFFFF}{\color[HTML]{000000} 63.59}} &
  {\color[HTML]{000000} 13.98} &
  {\color[HTML]{000000} 19.75} &
  {\color[HTML]{000000} 68.98} &
  {\color[HTML]{000000} 24.20} &
  \multicolumn{1}{c!{\vrule width 1.1pt}}{\cellcolor[HTML]{FFFFFF}{\color[HTML]{000000} 73.10}} &
  {\color[HTML]{000000} 18.29} &
  {\color[HTML]{000000} 22.23} &
  {\color[HTML]{000000} 76.24} &
  {\color[HTML]{000000} 28.79} &
  \multicolumn{1}{c!{\vrule width 1.1pt}}{\cellcolor[HTML]{FFFFFF}{\color[HTML]{000000} 81.49}} &
  {\color[HTML]{000000} 13.19} &
  {\color[HTML]{000000} 17.68} &
  {\color[HTML]{000000} 53.68} &
  {\color[HTML]{000000} 22.06} &
  \multicolumn{1}{c!{\vrule width 1.1pt}}{\cellcolor[HTML]{FFFFFF}{\color[HTML]{000000} 58.04}} &
  {\color[HTML]{000000} \textbf{10.60}} &
  {\color[HTML]{000000} \textbf{12.65}} &
  {\color[HTML]{000000} \textbf{33.06}} &
  {\color[HTML]{000000} \textbf{16.51}} &
  {\color[HTML]{000000} \textbf{36.95}} \\ \midrule[1.1pt]
\multicolumn{1}{c!{\vrule width 1.1pt}}{\cellcolor[HTML]{FFFFFF}{\color[HTML]{000000} }} &
  \multicolumn{1}{c!{\vrule width 1.1pt}}{\cellcolor[HTML]{FFFFFF}{\color[HTML]{000000} A}} &
  {\color[HTML]{000000} 0.86} &
  {\color[HTML]{000000} 0.69} &
  {\color[HTML]{000000} 3.80} &
  {\color[HTML]{000000} 1.10} &
  \multicolumn{1}{c!{\vrule width 1.1pt}}{\cellcolor[HTML]{FFFFFF}{\color[HTML]{000000} 3.95}} &
  {\color[HTML]{000000} 0.56} &
  {\color[HTML]{000000} 1.27} &
  {\color[HTML]{000000} 9.65} &
  {\color[HTML]{000000} 1.39} &
  \multicolumn{1}{c!{\vrule width 1.1pt}}{\cellcolor[HTML]{FFFFFF}{\color[HTML]{000000} 9.75}} &
  {\color[HTML]{000000} 1.50} &
  {\color[HTML]{000000} 1.58} &
  {\color[HTML]{000000} 5.02} &
  {\color[HTML]{000000} 2.18} &
  \multicolumn{1}{c!{\vrule width 1.1pt}}{\cellcolor[HTML]{FFFFFF}{\color[HTML]{000000} 5.47}} &
  {\color[HTML]{000000} 0.59} &
  {\color[HTML]{000000} 0.77} &
  {\color[HTML]{000000} 1.87} &
  {\color[HTML]{000000} 0.97} &
  \multicolumn{1}{c!{\vrule width 1.1pt}}{\cellcolor[HTML]{FFFFFF}{\color[HTML]{000000} 2.10}} &
  {\color[HTML]{000000} \textbf{0.44}} &
  {\color[HTML]{000000} \textbf{0.33}} &
  {\color[HTML]{000000} \textbf{1.29}} &
  {\color[HTML]{000000} \textbf{0.55}} &
  {\color[HTML]{000000} \textbf{1.41}} \\
\multicolumn{1}{c!{\vrule width 1.1pt}}{\cellcolor[HTML]{FFFFFF}{\color[HTML]{000000} }} &
  \multicolumn{1}{c!{\vrule width 1.1pt}}{\cellcolor[HTML]{FFFFFF}{\color[HTML]{000000} B}} &
  {\color[HTML]{000000} 1.25} &
  {\color[HTML]{000000} 1.23} &
  {\color[HTML]{000000} 3.99} &
  {\color[HTML]{000000} 1.75} &
  \multicolumn{1}{c!{\vrule width 1.1pt}}{\cellcolor[HTML]{FFFFFF}{\color[HTML]{000000} 4.36}} &
  {\color[HTML]{000000} 0.91} &
  {\color[HTML]{000000} 1.85} &
  {\color[HTML]{000000} 8.81} &
  {\color[HTML]{000000} 2.06} &
  \multicolumn{1}{c!{\vrule width 1.1pt}}{\cellcolor[HTML]{FFFFFF}{\color[HTML]{000000} 9.05}} &
  {\color[HTML]{000000} 2.08} &
  {\color[HTML]{000000} 2.35} &
  {\color[HTML]{000000} 6.44} &
  {\color[HTML]{000000} 3.14} &
  \multicolumn{1}{c!{\vrule width 1.1pt}}{\cellcolor[HTML]{FFFFFF}{\color[HTML]{000000} 7.16}} &
  {\color[HTML]{000000} 0.87} &
  {\color[HTML]{000000} 1.34} &
  {\color[HTML]{000000} 4.38} &
  {\color[HTML]{000000} 1.59} &
  \multicolumn{1}{c!{\vrule width 1.1pt}}{\cellcolor[HTML]{FFFFFF}{\color[HTML]{000000} 4.66}} &
  {\color[HTML]{000000} \textbf{0.65}} &
  {\color[HTML]{000000} \textbf{0.52}} &
  {\color[HTML]{000000} \textbf{1.73}} &
  {\color[HTML]{000000} \textbf{0.84}} &
  {\color[HTML]{000000} \textbf{1.92}} \\
\multicolumn{1}{c!{\vrule width 1.1pt}}{\multirow{-3}{*}{\cellcolor[HTML]{FFFFFF}{\color[HTML]{000000} GREAT}}} &
  \multicolumn{1}{c!{\vrule width 1.1pt}}{\cellcolor[HTML]{FFFFFF}{\color[HTML]{000000} C}} &
  {\color[HTML]{000000} 1.21} &
  {\color[HTML]{000000} 2.16} &
  {\color[HTML]{000000} 4.46} &
  {\color[HTML]{000000} 2.48} &
  \multicolumn{1}{c!{\vrule width 1.1pt}}{\cellcolor[HTML]{FFFFFF}{\color[HTML]{000000} 5.10}} &
  {\color[HTML]{000000} 0.84} &
  {\color[HTML]{000000} 1.68} &
  {\color[HTML]{000000} 10.28} &
  {\color[HTML]{000000} 1.88} &
  \multicolumn{1}{c!{\vrule width 1.1pt}}{\cellcolor[HTML]{FFFFFF}{\color[HTML]{000000} 10.45}} &
  {\color[HTML]{000000} 1.60} &
  {\color[HTML]{000000} 2.56} &
  {\color[HTML]{000000} 4.59} &
  {\color[HTML]{000000} 3.02} &
  \multicolumn{1}{c!{\vrule width 1.1pt}}{\cellcolor[HTML]{FFFFFF}{\color[HTML]{000000} 5.49}} &
  {\color[HTML]{000000} 1.25} &
  {\color[HTML]{000000} 2.17} &
  {\color[HTML]{000000} 2.81} &
  {\color[HTML]{000000} 2.50} &
  \multicolumn{1}{c!{\vrule width 1.1pt}}{\cellcolor[HTML]{FFFFFF}{\color[HTML]{000000} 3.76}} &
  {\color[HTML]{000000} \textbf{1.19}} &
  {\color[HTML]{000000} \textbf{1.33}} &
  {\color[HTML]{000000} \textbf{2.75}} &
  {\color[HTML]{000000} \textbf{1.79}} &
  {\color[HTML]{000000} \textbf{3.28}} \\ \midrule[1.1pt]
\multicolumn{1}{c!{\vrule width 1.1pt}}{\cellcolor[HTML]{FFFFFF}{\color[HTML]{000000} }} &
  \multicolumn{1}{c!{\vrule width 1.1pt}}{\cellcolor[HTML]{FFFFFF}{\color[HTML]{000000} A}} &
  {\color[HTML]{000000} 0.97} &
  {\color[HTML]{000000} 1.32} &
  {\color[HTML]{000000} 4.58} &
  {\color[HTML]{000000} 1.64} &
  \multicolumn{1}{c!{\vrule width 1.1pt}}{\cellcolor[HTML]{FFFFFF}{\color[HTML]{000000} 4.87}} &
  {\color[HTML]{000000} 2.15} &
  {\color[HTML]{000000} 2.32} &
  {\color[HTML]{000000} 13.84} &
  {\color[HTML]{000000} 3.16} &
  \multicolumn{1}{c!{\vrule width 1.1pt}}{\cellcolor[HTML]{FFFFFF}{\color[HTML]{000000} 14.20}} &
  {\color[HTML]{000000} 2.22} &
  {\color[HTML]{000000} 2.37} &
  {\color[HTML]{000000} 4.61} &
  {\color[HTML]{000000} 3.25} &
  \multicolumn{1}{c!{\vrule width 1.1pt}}{\cellcolor[HTML]{FFFFFF}{\color[HTML]{000000} 5.64}} &
  {\color[HTML]{000000} 2.06} &
  {\color[HTML]{000000} 1.47} &
  {\color[HTML]{000000} 5.72} &
  {\color[HTML]{000000} 2.53} &
  \multicolumn{1}{c!{\vrule width 1.1pt}}{\cellcolor[HTML]{FFFFFF}{\color[HTML]{000000} 6.25}} &
  {\color[HTML]{000000} \textbf{0.82}} &
  {\color[HTML]{000000} \textbf{1.03}} &
  {\color[HTML]{000000} \textbf{1.57}} &
  {\color[HTML]{000000} \textbf{1.32}} &
  {\color[HTML]{000000} \textbf{2.05}} \\
\multicolumn{1}{c!{\vrule width 1.1pt}}{\cellcolor[HTML]{FFFFFF}{\color[HTML]{000000} }} &
  \multicolumn{1}{c!{\vrule width 1.1pt}}{\cellcolor[HTML]{FFFFFF}{\color[HTML]{000000} B}} &
  {\color[HTML]{000000} 1.08} &
  {\color[HTML]{000000} 1.47} &
  {\color[HTML]{000000} 4.42} &
  {\color[HTML]{000000} 1.83} &
  \multicolumn{1}{c!{\vrule width 1.1pt}}{\cellcolor[HTML]{FFFFFF}{\color[HTML]{000000} 4.78}} &
  {\color[HTML]{000000} 1.69} &
  {\color[HTML]{000000} 2.40} &
  {\color[HTML]{000000} 12.32} &
  {\color[HTML]{000000} 2.94} &
  \multicolumn{1}{c!{\vrule width 1.1pt}}{\cellcolor[HTML]{FFFFFF}{\color[HTML]{000000} 12.66}} &
  {\color[HTML]{000000} 1.79} &
  {\color[HTML]{000000} 2.18} &
  {\color[HTML]{000000} 5.57} &
  {\color[HTML]{000000} 2.82} &
  \multicolumn{1}{c!{\vrule width 1.1pt}}{\cellcolor[HTML]{FFFFFF}{\color[HTML]{000000} 6.25}} &
  {\color[HTML]{000000} 1.76} &
  {\color[HTML]{000000} 1.84} &
  {\color[HTML]{000000} 9.31} &
  {\color[HTML]{000000} 2.54} &
  \multicolumn{1}{c!{\vrule width 1.1pt}}{\cellcolor[HTML]{FFFFFF}{\color[HTML]{000000} 9.65}} &
  {\color[HTML]{000000} \textbf{0.70}} &
  {\color[HTML]{000000} \textbf{0.49}} &
  {\color[HTML]{000000} \textbf{1.94}} &
  {\color[HTML]{000000} \textbf{0.86}} &
  {\color[HTML]{000000} \textbf{2.12}} \\
\multicolumn{1}{c!{\vrule width 1.1pt}}{\multirow{-3}{*}{\cellcolor[HTML]{FFFFFF}{\color[HTML]{000000} PLANET}}} &
  \multicolumn{1}{c!{\vrule width 1.1pt}}{\cellcolor[HTML]{FFFFFF}{\color[HTML]{000000} C}} &
  {\color[HTML]{000000} 2.92} &
  {\color[HTML]{000000} 3.71} &
  {\color[HTML]{000000} 7.03} &
  {\color[HTML]{000000} 4.72} &
  \multicolumn{1}{c!{\vrule width 1.1pt}}{\cellcolor[HTML]{FFFFFF}{\color[HTML]{000000} 8.47}} &
  {\color[HTML]{000000} 3.16} &
  {\color[HTML]{000000} 4.10} &
  {\color[HTML]{000000} 18.11} &
  {\color[HTML]{000000} 5.18} &
  \multicolumn{1}{c!{\vrule width 1.1pt}}{\cellcolor[HTML]{FFFFFF}{\color[HTML]{000000} 18.83}} &
  {\color[HTML]{000000} 3.00} &
  {\color[HTML]{000000} 3.75} &
  {\color[HTML]{000000} 17.07} &
  {\color[HTML]{000000} 4.80} &
  \multicolumn{1}{c!{\vrule width 1.1pt}}{\cellcolor[HTML]{FFFFFF}{\color[HTML]{000000} 17.74}} &
  {\color[HTML]{000000} 3.28} &
  {\color[HTML]{000000} 3.79} &
  {\color[HTML]{000000} 15.34} &
  {\color[HTML]{000000} 5.01} &
  \multicolumn{1}{c!{\vrule width 1.1pt}}{\cellcolor[HTML]{FFFFFF}{\color[HTML]{000000} 16.14}} &
  {\color[HTML]{000000} \textbf{1.73}} &
  {\color[HTML]{000000} \textbf{1.40}} &
  {\color[HTML]{000000} \textbf{5.47}} &
  {\color[HTML]{000000} \textbf{2.23}} &
  {\color[HTML]{000000} \textbf{5.90}} \\ \bottomrule[1.3pt]
\end{tabular}
\end{table*}

\subsubsection{\normalfont Network Architecture}  
To effectively model satellite signal characteristics and their dynamic uncertainties, we design a deep learning-based architecture that leverages self-attention, normalization, and fully connected layers. The network consists of the following key components:

\begin{itemize}
    \item Multi-head Attention: The input satellite features are processed by a multi-head self-attention layer with \( N_h = 4 \) attention heads. This mechanism enables the model to capture interactions between different satellite observations, allowing dynamic feature weighting. The attention mechanism operates on the input embedding with residual connections, which enhances gradient flow and stabilizes training. The attention scores are computed as:
    \begin{equation}
    \text{Attention}(\mathbf{Q}, \mathbf{K}, \mathbf{V}) = \text{softmax} \left( \frac{\mathbf{Q} \mathbf{K}^\top}{\sqrt{d_k}} \right) \mathbf{V}
    \end{equation}
    where \( \mathbf{Q}, \mathbf{K}, \mathbf{V} \) are the query, key, and value matrices derived from the input embeddings, and \( d_k \) is the scaling factor.

    \item Normalization Layer: To improve training stability and feature consistency, a normalization layer (e.g., Layer Normalization or Batch Normalization) is applied after the attention mechanism. This ensures that feature distributions remain stable across different samples, reducing covariate shift and improving convergence speed. The transformation is defined as:
    \begin{equation}
    \hat{\mathbf{X}} = \frac{\mathbf{X} - \mu}{\sigma + \epsilon} \gamma + \beta
    \end{equation}
    where \( \mu \) and \( \sigma \) denote the mean and standard deviation, respectively, computed across the feature dimension. \( \gamma \) and \( \beta \) are trainable scaling and shifting parameters, and \( \epsilon \) is a small constant for numerical stability.

    \item Fully Connected Layers (MLP): The model contains three fully connected layers with hidden dimensions of 64, 128, and 64, respectively. These layers apply nonlinear transformations to the normalized attention-refined embeddings, further enhancing feature extraction. Specifically, the transformation at each layer is given by:
    \begin{equation}
    \mathbf{h}_{i+1} = \text{ReLU} (\mathbf{W}_i \mathbf{h}_i + \mathbf{b}_i)
    \end{equation}
    where \( \mathbf{W}_i \) and \( \mathbf{b}_i \) are the weight and bias parameters of the \( i \)-th layer.

    \item Output Layer: The final fully connected layer maps the learned features to a 2-dimensional output:
    \begin{itemize}
        \item \( \mathbf{R}_{\text{diag}} \): The diagonal elements of the measurement noise covariance matrix, estimated using the \texttt{Softplus} activation function to ensure non-negativity.
        \item \( \mathbf{v}^{\mathbf{c}} \): The innovation compensation terms for each satellite observation, directly predicted as real-valued outputs.
    \end{itemize}
\end{itemize}

\subsubsection{\normalfont Forward Pass}  
The forward pass of the network follows these steps:  
\begin{itemize}
    \item Attention Processing: The input feature matrix \( \mathbf{X} \in \mathbb{R}^{B \times N \times d} \), where \( B \) is the batch size, \( N \) is the number of satellites, and \( d \) is the feature dimension (set to \( d = 8 \) in LF-GNSS), is processed by the multi-head self-attention mechanism with 4 attention heads, capturing dependencies between satellites.

    \item Residual Connection: A skip connection is employed to combine the original input with the attention-refined features, forming a residual pathway that improves gradient flow and enhances model stability.

    \item Normalization Layer: The attention-enhanced feature representations are normalized to stabilize feature distributions before being passed into fully connected layers.

    \item Layer-wise Processing: The normalized features are sequentially passed through three fully connected layers with dimensions \( 64 \to 128 \to 64 \), each followed by a ReLU activation to introduce nonlinearities and refine the learned representations.

    \item Output Computation: The outputs from the final layer are defined as:
    \begin{equation}
    \mathbf{R}_{\text{diag}} = \text{Softplus}(\mathbf{o}_{R}), \quad \mathbf{v}^{\mathbf{c}} = \mathbf{o}_{v}
    \end{equation}
    where \( \mathbf{o}_{R} \) and \( \mathbf{o}_{v} \) represent the components of the output matrix \( \mathbf{o} \), corresponding to the measurement noise covariance and the innovation compensation vector, respectively. The Softplus function is used to ensure that the predicted noise covariance remains non-negative:
    \begin{equation}
    \text{Softplus}(x) = \ln(1 + e^{x})
    \end{equation}
\end{itemize}

\subsection{Extended Kalman Filter Processor}  
Based on GNSS model, a EKF processor is constructed, with the state vector defined as:
\begin{equation}
\mathbf{x} = \left[ \mathbf{p}, \mathbf{v}, \Delta t_r, \dot{\Delta t_r}, \mathbf{ISB} \right]^\top
\end{equation}
This equation builds upon the state vector defined in equation \ref{state vector 1}, with the addition of the velocity and clock drift terms.

\subsubsection{Initialization}  
The state vector and covariance matrix are initialized as follows: The initial position \( \mathbf{p}_0 \) and clock biases \( \Delta t_r^0 \) of the receiver are obtained from the coarse positioning module, and all other parameters are initialized to zero.

The initial covariance matrix \( \mathbf{P}_0 \) is defined as a block diagonal matrix:  
\begin{equation}
    \mathbf{P}_0 = \text{diag}(
    \sigma_{p0}^2 \mathbf{I}_3, \,
    \sigma_{v0}^2 \mathbf{I}_3, \,
    \sigma_{cb0}^2 \mathbf{I}_1, \,
    \sigma_{cd0}^2 \mathbf{I}_1, \,
    \sigma_{ISB0}^2 \mathbf{I}_3)
\end{equation}
where \( \sigma^2 \) represents the uncertainties of the state variables.

\subsubsection{Time Update}  
In the LF-GNSS framework, a constant velocity model is adopted to describe the dynamic behavior of the receiver. The changes in clock drift and ISB are modeled as  zero mean Gaussian white noise. Under this assumption, the state transition matrix \( \mathbf{F}_k \) can be expressed as:
\begin{equation}
    \mathbf{F}_k = 
    \begin{bmatrix}
    \mathbf{I}_3 & \mathbf{I}_3 \Delta t & \mathbf{0}_{3 \times 1} & \mathbf{0}_{3 \times 1} & \mathbf{0}_{3 \times 3} \\
    \mathbf{0}_{3 \times 3} & \mathbf{I}_3 & \mathbf{0}_{3 \times 1} & \mathbf{0}_{3 \times 1} & \mathbf{0}_{3 \times 3} \\
    \mathbf{0}_{1 \times 3} & \mathbf{0}_{1 \times 3} & \mathbf{I}_1 & \mathbf{I}_1 \Delta t & \mathbf{0}_{1 \times 3} \\
    \mathbf{0}_{1 \times 3} & \mathbf{0}_{1 \times 3} & \mathbf{0}_{1 \times 1} & \mathbf{I}_1 & \mathbf{0}_{1 \times 3} \\
    \mathbf{0}_{3 \times 3} & \mathbf{0}_{3 \times 3} & \mathbf{0}_{3 \times 1} & \mathbf{0}_{3 \times 1} & \mathbf{I}_3
    \end{bmatrix}
\end{equation}
where \( \Delta t \) is the time interval between epochs.

The time update equations for the state vector and covariance matrix are given by:  
\begin{equation}
    \mathbf{x}_{k|k-1} = \mathbf{F}_k \mathbf{x}_{k-1} + \mathbf{u}_k
\end{equation}
\begin{equation}
    \mathbf{P}_{k|k-1} = \mathbf{F}_k \mathbf{P}_{k-1|k-1} \mathbf{F}_k^\top + \mathbf{Q}_k
\end{equation}
where \( \mathbf{u}_k \) is the process noise vector, and \( \mathbf{Q}_k \) represents the process noise covariance matrix.

\subsubsection{Measurement Update}  
The measurement update process refines the predicted state by incorporating incoming GNSS measurements. This involves the following steps:  
\begin{itemize}
    \item Compute Innovation Vector:  
    \begin{equation}
        \mathbf{v}_k = \mathbf{z}_k - \mathbf{H}_k \mathbf{x}_{k|k-1}
    \end{equation}
    where \( \mathbf{z}_k \), the measurement vector, represents the pseudorange observations from visible satellites, and \( \mathbf{H}_k \) maps these measurements to the state vector. Their detailed definitions are provided in Equations \ref{z} and \ref{H}.
    \item Compute Innovation Covariance:  
    \begin{equation}
        \mathbf{S}_k = \mathbf{H}_k \mathbf{P}_{k|k-1} \mathbf{H}_k^\top + \mathbf{R}_k
    \end{equation}
    In our framework, \( \mathbf{R}_k \) is directly estimated by the DL network. This matrix is modeled as a diagonal matrix, representing independent measurement noise for each observed satellite:  
    \begin{equation}
        \mathbf{R}_k = \text{diag}(r_1, r_2, \dots, r_N)
    \end{equation}
    where \( N \) is the number of visible satellites at epoch \( k \), and each diagonal element \( r_n \) represents the estimated measurement noise variance for the \( n \)-th satellite. LF-GNSS can dynamically adapt measurement noise levels based on satellite conditions, improving the robustness and accuracy of the state estimation process.

    \item Compute Kalman Gain:  
    \begin{equation}
        \mathbf{K}_k = \mathbf{P}_{k|k-1} \mathbf{H}_k^\top \mathbf{S}_k^{-1}
    \end{equation}

    \item Update State Estimate:  
    \begin{equation}
        \mathbf{x}_{k} = \mathbf{x}_{k|k-1} + \mathbf{K}_k (\mathbf{v}_k + \mathbf{v}_k^{\mathbf{c}})
    \end{equation}
    where \( \mathbf{v}_k^{\mathbf{c}} \) denotes the innovation compensation term, which is directly predicted by the DL network. This compensation term enhances the state update by mitigating the impact of unmodeled errors.

    \item Update Covariance Matrix:  
    \begin{equation}
        \mathbf{P}_{k} = (\mathbf{I} - \mathbf{K}_k \mathbf{H}_k) \mathbf{P}_{k|k-1} (\mathbf{I} - \mathbf{K}_k \mathbf{H}_k)^\top + \mathbf{K}_k \mathbf{R}_k \mathbf{K}_k^\top
    \end{equation}
    The Joseph stabilized update formula is applied to ensure numerical stability in LF-GNSS.
\end{itemize}

\begin{figure*}[!t]
\centerline{\includegraphics[width=180mm]{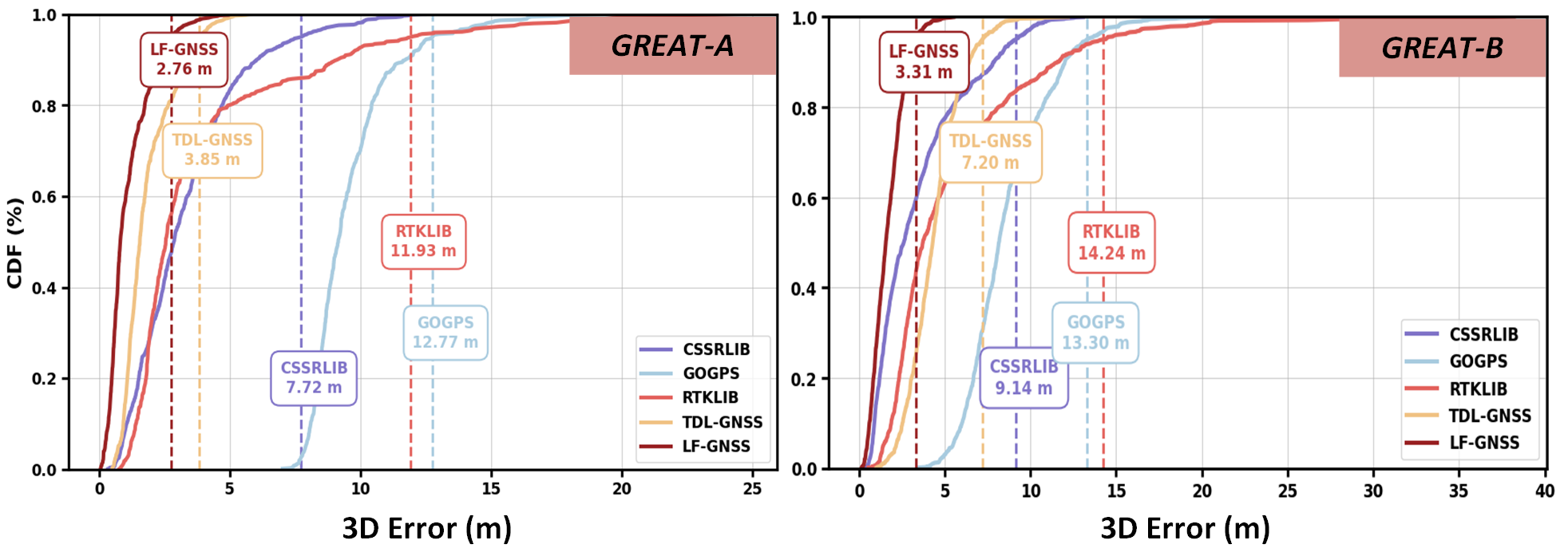}}
\caption{Public experiment results: CDF analysis of 3D positioning errors across all frameworks for datasets GREAT-A and GREAT-B.}
\label{Fig3_ErrorCDFs}
\end{figure*}

\begin{figure*}[!t]
\centering
\includegraphics[width=\linewidth,height=90mm]{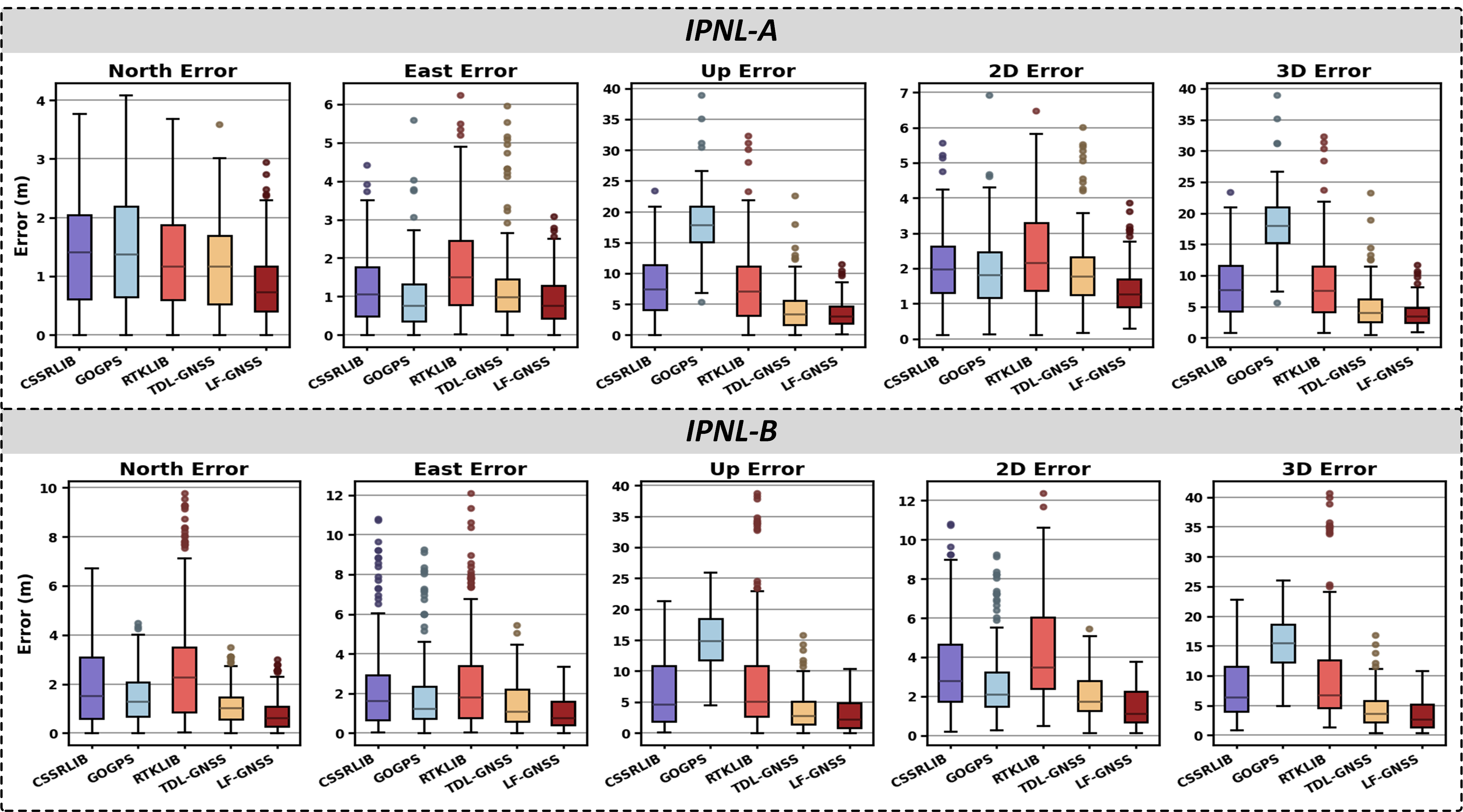}
\caption{Public experiment results: boxplot analysis of positioning errors across all frameworks for datasets IPNL-A and IPNL-B.}
\label{Fig4_ErrorBox}
\end{figure*}

\subsection{Dynamic Hard Example Mining}
To address the imbalance between easy and hard samples in regression tasks, we adapt the focal loss mechanism, originally designed for classification, and extend it to handle continuous regression errors. Our approach introduces a dynamic weighting factor based on the Euclidean distance between predictions and targets, allowing the model to focus more on hard samples.

First, we extract the three-dimensional positioning results from the state estimation output, which are processed through the EKF processor. Let \( \mathbf{x}_{\text{pred}} \in \mathbb{R}^3 \) denote the predicted ENU coordinates and \( \mathbf{x}_{\text{true}} \in \mathbb{R}^3 \) denote the ground truth ENU coordinates. The ENU error between the predicted and true coordinates is computed as:
\begin{equation}
\mathbf{e}_{\text{ENU}} = \mathbf{x}_{\text{pred}} - \mathbf{x}_{\text{true}}
\end{equation}
where \( \mathbf{x}_{\text{pred}} = [e_{\text{pred}}, n_{\text{pred}}, u_{\text{pred}}] \) and \( \mathbf{x}_{\text{true}} = [e_{\text{true}}, n_{\text{true}}, u_{\text{true}}] \) are the predicted and true ENU coordinates, respectively.


Next, we compute the Euclidean distance of the ENU error, which also represents the base loss:
\begin{equation}
L_{\text{base}} = d = \|\mathbf{e}_{\text{ENU}}\|_2 = \sqrt{\sum_{i=1}^{3} e_{i,\text{ENU}}^2}
\end{equation}
where \( e_{i,\text{ENU}} \) represents the components of the ENU error, with \( i \in \{1, 2, 3\} \), corresponding to the \( e \), \( n \), and \( u \) directions of the error.

To focus on hard examples, we introduce a weighting factor in the loss function. This factor dynamically adjusts the importance of each sample based on the Euclidean distance. The weighting factor \( w \) is defined as:
\begin{equation}
w = \left( 1 - \sqrt{\frac{L_{\text{base}}}{\max(L_{\text{base}})}} \right)^{\gamma_{\text{dynamic}}}
\end{equation}
where \( \max(L_{\text{base}}) \) is the maximum error in the batch, and \( \gamma_{\text{dynamic}} \) is a dynamic parameter that adjusts the weight based on the difficulty of the sample.

To further adjust the influence of hard examples, we define a dynamic \( \gamma \) parameter that scales with the base loss. If dynamic adjustment is enabled, \( \gamma_{\text{dynamic}} \) is computed as:
\begin{equation}
\gamma_{\text{dynamic}} = \gamma \cdot \exp\left(-\lambda \cdot L_{\text{base}}\right)
\end{equation}
where \( \gamma \) is the initial difficulty weight, and \( \lambda \) controls the rate of change of \( \gamma_{\text{dynamic}} \).

The final dynamic hard example mining loss, \( L_{\text{dhem}} \), is computed as the product of the base loss and the weighting factor \( w \), scaled by a hyperparameter \( \alpha \) to control the overall loss magnitude:
\begin{equation}
L_{\text{dhem}} = \alpha \cdot w \cdot L_{\text{base}}
\end{equation}

Finally, to ensure that the loss is a stable scalar, we compute the root mean square error (RMSE) form of the loss:
\begin{equation}
L_{\text{dhem}}^{\text{RMSE}} = \sqrt{\frac{1}{N} \sum_{i=1}^{N} \left(L_{\text{dhem}}^{(i)}\right)^2}
\end{equation}
where \( N \) is the number of samples in the batch, and \( L_{\text{dhem}}^{(i)} \) is the dynamic hard example mining loss for the \( i \)-th sample.

By leveraging a dynamic hard example mining loss, the model prioritizes samples with larger errors during training, enhancing its ability to handle challenging cases and significantly improving performance in high-error scenarios.
\begin{figure}[!t]
\centering
\includegraphics[width=\linewidth]{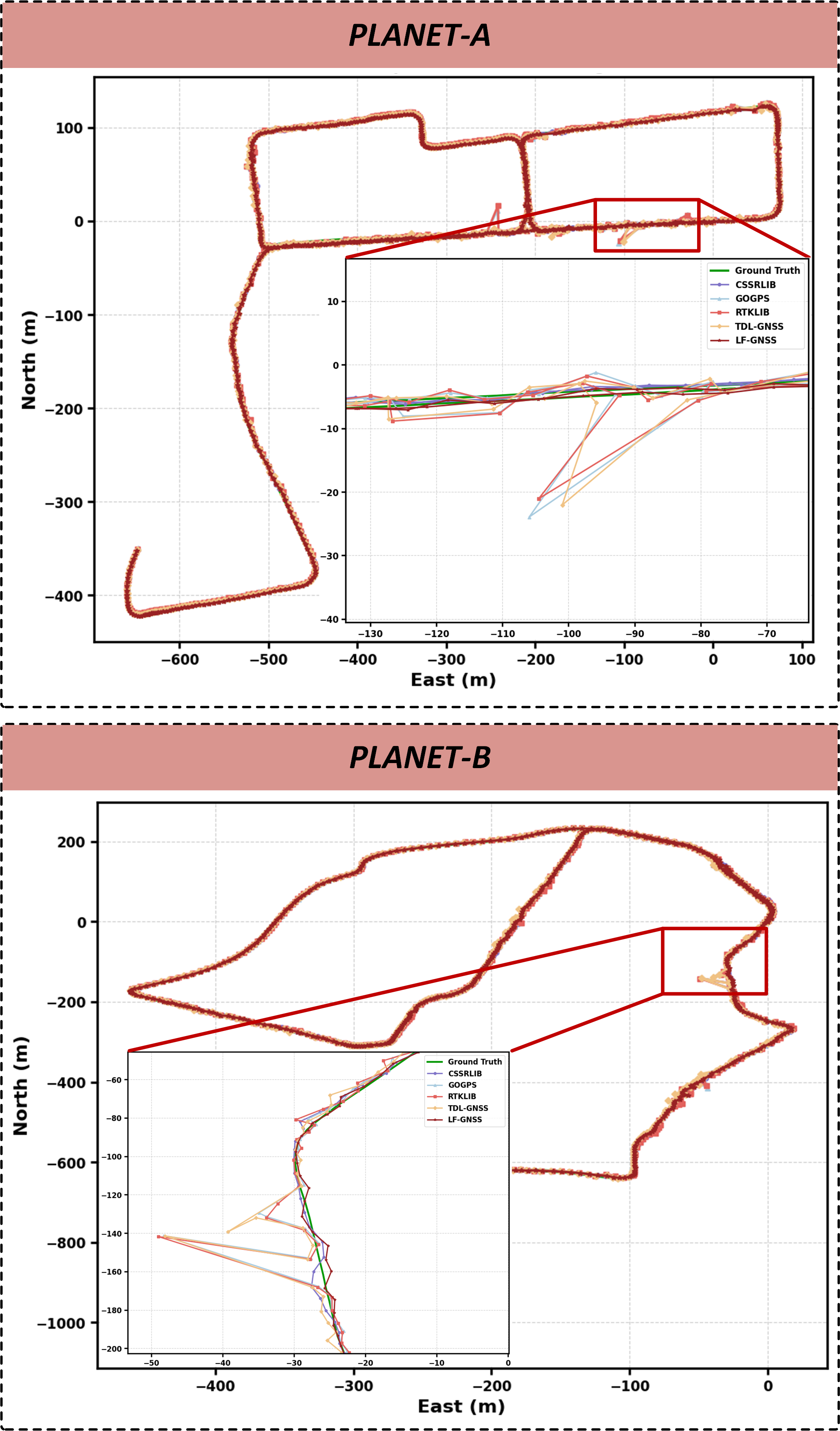}
\caption{Public experiment results: trajectory comparisons across all frameworks for datasets PLANET-A and PLANET-B.}
\label{Fig5_Trajs}
\end{figure}

\section{Experimental validation}
\subsection{Experimental Setup and Details}

We test the proposed LF-GNSS system both on public datasets and real-world experiments. The positioning error metrics used to assess the performance of LF-GNSS are computed in the ENU coordinate system and evaluated using the RMSE. The experiments were conducted on a system running Ubuntu 20.04, equipped with an Intel i9-14900KF CPU and a NVIDIA GeForce RTX 4090 GPU. The model was trained for 200 epochs using the Adam optimizer with a batch size of 16 and an initial learning rate of 0.001. A CosineAnnealingLR scheduler was applied, with \( T_{max} = 50 \) and \( \eta_{min} = 1 \times 10^{-4} \).

\subsection{Public Dataset Experiments}
We extensively selected the KLT-[3, 1, 2] \cite{hu2023fisheye}, and Whampoa datasets from the IPNL research group, which are referred to in this paper as IPNL-Train, IPNL-A, IPNL-B, and IPNL-C, respectively. Additionally, we used the Campus-[01, 02, 03] and Campus-night datasets \cite{great2024dataset} from the GREAT research group, which are referred to as GREAT-Train, GREAT-A, GREAT-B, and GREAT-C in this paper. Finally, we selected a long-duration and complex dataset from the SmartPNT-POS \cite{zhu2024large} released by the PLANET research team . This dataset was divided into four subsets for our experiments: PLANET-Train, PLANET-A, PLANET-B, and PLANET-C. In each dataset, the "-Train" subset was specifically used for training purposes, while the remaining subsets were used for testing and evaluation in the experiments.

We compared LF-GNSS with the DL-based framework TDL-GNSS, as well as traditional open-source frameworks including CSSRLIB, RTKLIB, and goGPS. The detailed results are provided in Table \ref{Public Dataset Experiments Results}. The comparison highlights that our framework achieves superior positioning performance.

Fig. \ref{Fig3_ErrorCDFs} presents the Cumulative Distribution Functions (CDFs) for datasets GREAT-A and GREAT-B. The LF-GNSS curve demonstrates a steeper rise, reaching a cumulative probability of 1.0 more quickly. Additionally, the 95\% error statistics confirm the advantage of LF-GNSS in positioning accuracy compared to other frameworks.

Fig. \ref{Fig4_ErrorBox} illustrates the error distribution for datasets IPNL-A and IPNL-B using boxplots. The results show that LF-GNSS achieves a more concentrated error distribution around zero, with fewer outliers compared to other methods. This highlights the framework's superior positioning accuracy and stability.

Fig. \ref{Fig5_Trajs} compares the trajectories for datasets PLANET-A and PLANET-B. The trajectories generated by LF-GNSS align more closely with the ground truth, showcasing its enhanced accuracy and reliability in challenging scenarios.

\begin{figure*}[!t]
\centering
\includegraphics[width=\linewidth,height=80mm]{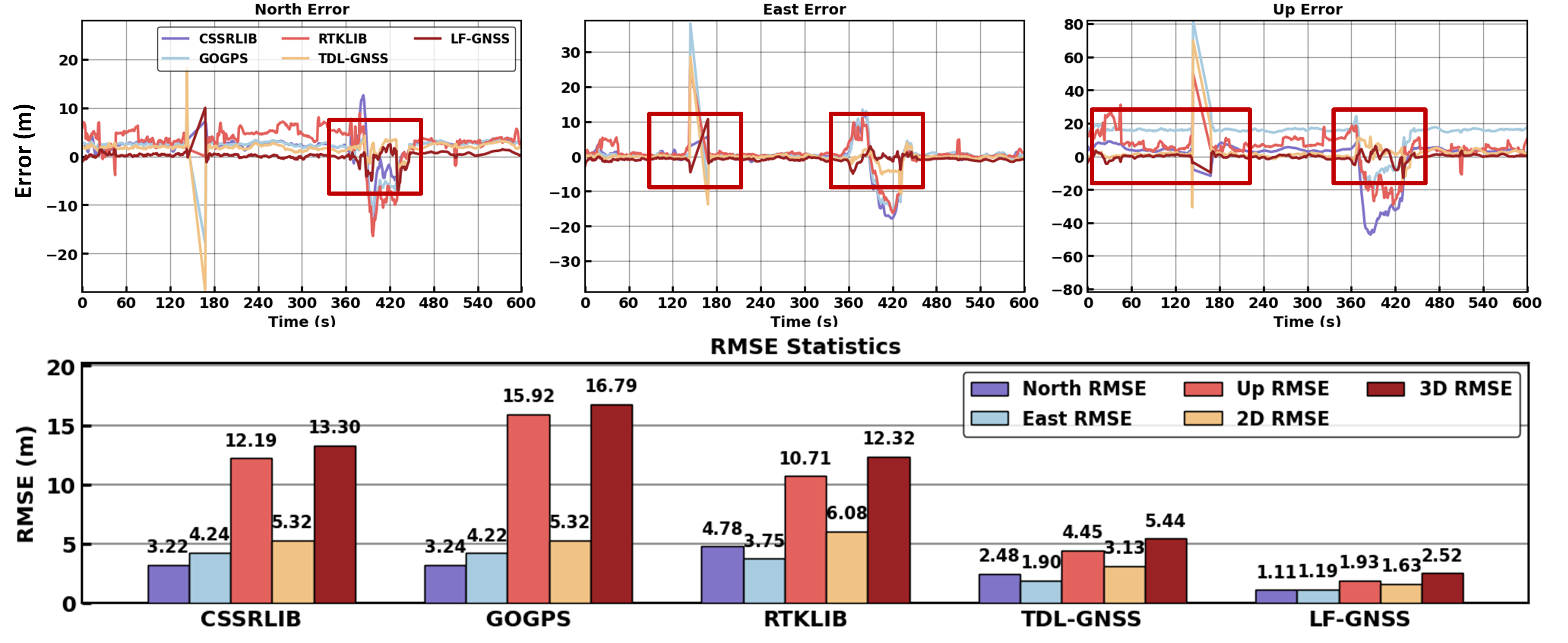}
\caption{Private experiment results: E/N/U error curves and RMSE analysis across frameworks for dataset ALGO-C.}
\label{Fig7_Errors_RMSE}
\end{figure*}

\begin{table*}[]
\centering
\caption{Private Dataset Expriment Results [RMSE, m]}
\label{Private Dataset Experiments Results}
\renewcommand{\arraystretch}{1.3} 
\setlength{\tabcolsep}{8pt} 
\begin{tabular}{cclclclclclclclcl
>{\columncolor[HTML]{F7E0DF}}c 
>{\columncolor[HTML]{F7E0DF}}l }
\toprule[1.3pt]
\multicolumn{3}{c}{\textbf{Method}} &
  \multicolumn{2}{c}{\textbf{CSSRLIB}} &
  \multicolumn{2}{c}{\textbf{goGPS}} &
  \multicolumn{2}{c}{\textbf{RTKLIB}} &
  \multicolumn{2}{c}{\textbf{TDL-GNSS}} &
  \multicolumn{2}{c}{\textbf{\begin{tabular}[c]{@{}c@{}}LF-GNSS\\ (w/o DPC+HEM)\end{tabular}}} &
  \multicolumn{2}{c}{\textbf{\begin{tabular}[c]{@{}c@{}}LF-GNSS\\ (w/o DPC)\end{tabular}}} &
  \multicolumn{2}{c}{\textbf{\begin{tabular}[c]{@{}c@{}}LF-GNSS\\ (w/o HEM)\end{tabular}}} &
  \multicolumn{2}{c}{\cellcolor[HTML]{F7E0DF}\textbf{\begin{tabular}[c]{@{}c@{}}LF-GNSS\\ (Ours)\end{tabular}}} \\ \midrule[1.1pt]
\multicolumn{1}{c!{\vrule width 1.1pt}}{} &
  \multicolumn{2}{c!{\vrule width 1.1pt}}{2D} &
  \multicolumn{2}{c!{\vrule width 1.1pt}}{5.99} &
  \multicolumn{2}{c!{\vrule width 1.1pt}}{4.12} &
  \multicolumn{2}{c!{\vrule width 1.1pt}}{7.34} &
  \multicolumn{2}{c!{\vrule width 1.1pt}}{3.05} &
  \multicolumn{2}{c!{\vrule width 1.1pt}}{2.78} &
  \multicolumn{2}{c!{\vrule width 1.1pt}}{2.45} &
  \multicolumn{2}{c!{\vrule width 1.1pt}}{2.53} &
  \multicolumn{2}{c}{\cellcolor[HTML]{F7E0DF}\textbf{1.83}} \\
\multicolumn{1}{c!{\vrule width 1.1pt}}{\multirow{-2}{*}{\begin{tabular}[c]{@{}c@{}}ALGO\\ (A)\end{tabular}}} &
  \multicolumn{2}{c!{\vrule width 1.1pt}}{3D} &
  \multicolumn{2}{c!{\vrule width 1.1pt}}{19.84} &
  \multicolumn{2}{c!{\vrule width 1.1pt}}{24.52} &
  \multicolumn{2}{c!{\vrule width 1.1pt}}{27.73} &
  \multicolumn{2}{c!{\vrule width 1.1pt}}{5.20} &
  \multicolumn{2}{c!{\vrule width 1.1pt}}{4.99} &
  \multicolumn{2}{c!{\vrule width 1.1pt}}{4.89} &
  \multicolumn{2}{c!{\vrule width 1.1pt}}{5.02} &
  \multicolumn{2}{c}{\cellcolor[HTML]{F7E0DF}\textbf{4.87}} \\ \midrule[1.1pt]
\multicolumn{1}{c!{\vrule width 1.1pt}}{} &
  \multicolumn{2}{c!{\vrule width 1.1pt}}{2D} &
  \multicolumn{2}{c!{\vrule width 1.1pt}}{2.95} &
  \multicolumn{2}{c!{\vrule width 1.1pt}}{2.67} &
  \multicolumn{2}{c!{\vrule width 1.1pt}}{5.20} &
  \multicolumn{2}{c!{\vrule width 1.1pt}}{1.99} &
  \multicolumn{2}{c!{\vrule width 1.1pt}}{1.71} &
  \multicolumn{2}{c!{\vrule width 1.1pt}}{1.19} &
  \multicolumn{2}{c!{\vrule width 1.1pt}}{1.57} &
  \multicolumn{2}{c}{\cellcolor[HTML]{F7E0DF}\textbf{0.81}} \\
\multicolumn{1}{c!{\vrule width 1.1pt}}{\multirow{-2}{*}{\begin{tabular}[c]{@{}c@{}}ALGO\\ (B)\end{tabular}}} &
  \multicolumn{2}{c!{\vrule width 1.1pt}}{3D} &
  \multicolumn{2}{c!{\vrule width 1.1pt}}{6.17} &
  \multicolumn{2}{c!{\vrule width 1.1pt}}{17.62} &
  \multicolumn{2}{c!{\vrule width 1.1pt}}{10.57} &
  \multicolumn{2}{c!{\vrule width 1.1pt}}{2.70} &
  \multicolumn{2}{c!{\vrule width 1.1pt}}{2.45} &
  \multicolumn{2}{c!{\vrule width 1.1pt}}{2.12} &
  \multicolumn{2}{c!{\vrule width 1.1pt}}{2.34} &
  \multicolumn{2}{c}{\cellcolor[HTML]{F7E0DF}\textbf{1.91}} \\ \midrule[1.1pt]
\multicolumn{1}{c!{\vrule width 1.1pt}}{} &
  \multicolumn{2}{c!{\vrule width 1.1pt}}{2D} &
  \multicolumn{2}{c!{\vrule width 1.1pt}}{5.32} &
  \multicolumn{2}{c!{\vrule width 1.1pt}}{5.32} &
  \multicolumn{2}{c!{\vrule width 1.1pt}}{6.08} &
  \multicolumn{2}{c!{\vrule width 1.1pt}}{3.13} &
  \multicolumn{2}{c!{\vrule width 1.1pt}}{2.47} &
  \multicolumn{2}{c!{\vrule width 1.1pt}}{1.88} &
  \multicolumn{2}{c!{\vrule width 1.1pt}}{2.21} &
  \multicolumn{2}{c}{\cellcolor[HTML]{F7E0DF}\textbf{1.63}} \\
\multicolumn{1}{c!{\vrule width 1.1pt}}{\multirow{-2}{*}{\begin{tabular}[c]{@{}c@{}}ALGO\\ (C)\end{tabular}}} &
  \multicolumn{2}{c!{\vrule width 1.1pt}}{3D} &
  \multicolumn{2}{c!{\vrule width 1.1pt}}{13.30} &
  \multicolumn{2}{c!{\vrule width 1.1pt}}{16.79} &
  \multicolumn{2}{c!{\vrule width 1.1pt}}{12.32} &
  \multicolumn{2}{c!{\vrule width 1.1pt}}{5.44} &
  \multicolumn{2}{c!{\vrule width 1.1pt}}{4.86} &
  \multicolumn{2}{c!{\vrule width 1.1pt}}{3.79} &
  \multicolumn{2}{c!{\vrule width 1.1pt}}{4.59} &
  \multicolumn{2}{c}{\cellcolor[HTML]{F7E0DF}\textbf{2.52}} \\ \bottomrule[1.3pt]
\end{tabular}
\end{table*}

\begin{figure}[!t]
\centering
\includegraphics[width=\linewidth]{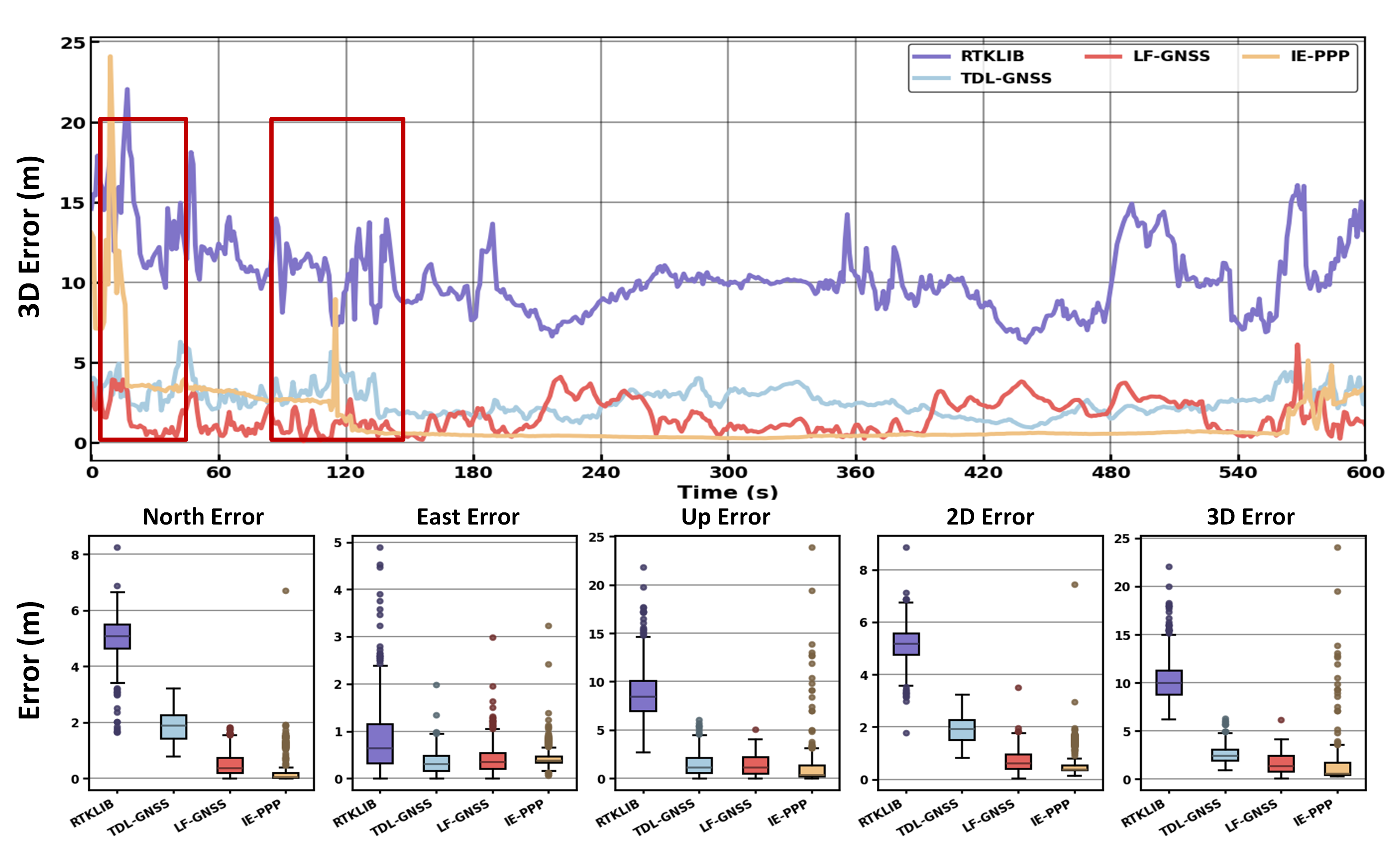}
\caption{Performance comparison with PPP: LF-GNSS vs IE-PPP on dataset ALGO-B.}
\label{Fig8_Errors_withPPP}
\end{figure}
\subsection{Private Dataset Experiments}
In the experiment, we set up an in-vehicle platform, as shown in Fig. \ref{Fig6_PrivateExperimentDetails}(a) and Fig. \ref{Fig6_PrivateExperimentDetails}(b). The platform was equipped with the Algo T1-3 device, capable of collecting raw GNSS and INS data and supporting Real-Time Kinematic (RTK)/INS Tightly Coupled (TC) processing. Raw GNSS observations were used as input for the evaluated algorithms, with RTK/INS TC solutions serving as ground truth. Fig. \ref{Fig6_PrivateExperimentDetails}(c) shows the test route through diverse environments, while Fig. \ref{Fig6_PrivateExperimentDetails}(d) highlights specific scenarios such as open-sky areas, tree-lined roads, urban canyons, and elevated bridges. These challenging scenarios provide a thorough assessment of algorithm performance.

We divided the 1.5-hour experimental dataset into four subsets: ALGO-Train, ALGO-A, ALGO-B, and ALGO-C. The ALGO-Train subset was used for training, while the remaining three subsets were designated for testing. The experimental results, presented in Table \ref{Private Dataset Experiments Results}, demonstrate that the proposed LF-GNSS framework consistently outperforms existing methods across all test datasets, achieving lower positioning errors.

The positioning results on dataset ALGO-C are illustrated in Fig. \ref{Fig7_Errors_RMSE}. It can be observed that LF-GNSS consistently maintains stable and reliable positioning performance even under challenging conditions, showcasing its robustness and adaptability. Notably, there is a significant improvement in stability within the red-boxed region, highlighting the model’s effectiveness in handling adverse scenarios. The final statistical analysis reveals an impressive 79.55\%–84.99\% improvement over traditional methods and a substantial 53.68\% enhancement compared to DL-based approaches, further demonstrating the superiority of the proposed framework.

\subsection{Ablation Study}
We conducted ablation studies to assess the effectiveness of the proposed DOP-based feature representation (DPC) and the Dynamic Example Mining method (HEM). The results, as shown in Table \ref{Private Dataset Experiments Results}, demonstrate the significant contribution of each innovation to the improvement in positioning accuracy. Specifically, the removal of the DPC feature (LF-GNSS w/o DPC) led to a noticeable deterioration in both 2D and 3D positioning accuracy compared to the full LF-GNSS framework, as evidenced by the increase in RMSE. Similarly, excluding the HEM method (LF-GNSS w/o HEM) also resulted in higher errors, particularly in urban environments, where the model faced more challenging satellite signals. These findings validate the importance of both the DPC feature and the HEM method in enhancing the model’s robustness and accuracy.

\subsection{Try to Compare with PPP}
PPP is a high-precision satellite positioning method that leverages carrier-phase measurements to achieve centimeter-level accuracy. In this study, we utilized Inertial Explorer (IE) software to compute PPP solutions for comparison.

As shown in Fig. \ref{Fig8_Errors_withPPP}, PPP struggles in complex urban environments, where carrier-phase outliers lead to significant fluctuations in positioning accuracy, particularly in the red-boxed areas. In contrast, LF-GNSS demonstrates consistently stable accuracy throughout the test. Statistical results further highlight its robustness, with LF-GNSS achieving a 3D RMSE of 1.91 m compared to 2.55 m for IE-PPP, indicating superior overall positioning accuracy across the entire dataset.

\section{Conclusion}
This paper presents LF-GNSS, a learning-filtering deep fusion framework designed to enhance satellite positioning in challenging urban environments. By leveraging deep learning networks, our framework dynamically analyzes satellite signals to adaptively construct observation noise covariance and compensated innovation vectors for Kalman filter input, mitigating NLOS and multipath effects. Additionally, we introduce a novel feature representation named DPC, which helps to better characterize the signal quality of individual satellites and improve measurement weighting. A dynamic hard example mining strategy further strengthens the model by prioritizing difficult signals during training. Ablation studies conducted to evaluate the effectiveness of the DPC feature and the HEM method demonstrate the substantial contributions of these innovations to overall performance improvement. Experimental results on both public and private datasets demonstrate that LF-GNSS consistently outperforms traditional and DL-based methods in positioning accuracy.

\bibliographystyle{IEEEtran}
\bibliography{ref}

\begin{thebibliography}{10}
\providecommand{\url}[1]{#1}
\csname url@samestyle\endcsname
\providecommand{\newblock}{\relax}
\providecommand{\bibinfo}[2]{#2}
\providecommand{\BIBentrySTDinterwordspacing}{\spaceskip=0pt\relax}
\providecommand{\BIBentryALTinterwordstretchfactor}{4}
\providecommand{\BIBentryALTinterwordspacing}{\spaceskip=\fontdimen2\font plus
\BIBentryALTinterwordstretchfactor\fontdimen3\font minus \fontdimen4\font\relax}
\providecommand{\BIBforeignlanguage}[2]{{%
\expandafter\ifx\csname l@#1\endcsname\relax
\typeout{** WARNING: IEEEtran.bst: No hyphenation pattern has been}%
\typeout{** loaded for the language `#1'. Using the pattern for}%
\typeout{** the default language instead.}%
\else
\language=\csname l@#1\endcsname
\fi
#2}}
\providecommand{\BIBdecl}{\relax}
\BIBdecl

\bibitem{zhang2024gnss}
H.~Zhang, C.-C. Chen, H.~Vallery, and T.~D. Barfoot, ``Gnss/multi-sensor fusion using continuous-time factor graph optimization for robust localization,'' \emph{IEEE Transactions on Robotics}, 2024.

\bibitem{groves2013portfolio}
P.~D. Groves, Z.~Jiang, M.~Rudi, and P.~Strode, ``A portfolio approach to nlos and multipath mitigation in dense urban areas.''\hskip 1em plus 0.5em minus 0.4em\relax The Institute of Navigation, 2013.

\bibitem{li2024enhanced}
B.~Li, L.~He, T.~Liu \emph{et~al.}, ``Enhanced land vehicular gnss/ins combined system by using multiple-antenna with common clock,'' \emph{IEEE Transactions on Vehicular Technology}, 2024.

\bibitem{shen2024novel}
Z.~Shen, X.~Li, X.~Wang, Z.~Wu, X.~Li, Y.~Zhou, and S.~Li, ``A novel factor graph framework for tightly coupled gnss/ins integration with carrier-phase ambiguity resolution,'' \emph{IEEE Transactions on Intelligent Transportation Systems}, 2024.

\bibitem{geng2024spoofing}
X.~Geng, Y.~Guo, K.~Tang, and W.~Wu, ``A spoofing algorithm for ground unmanned platform equipped with gnss/ins integrated navigation system,'' \emph{IEEE Transactions on Instrumentation and Measurement}, 2024.

\bibitem{geng2024covert}
X.~Geng, Y.~Guo, K.~Tang, W.~Wu, Y.~Ren, and G.~Duan, ``A covert spoofing algorithm for sins/gnss tightly integrated navigation system,'' \emph{IEEE Transactions on Automation Science and Engineering}, 2024.

\bibitem{song2024r2}
J.~Song, W.~Li, C.~Duan, and X.~Zhu, ``R2-gvio: A robust, real-time gnss-visual-inertial state estimator in urban challenging environments,'' \emph{IEEE Internet of Things Journal}, 2024.

\bibitem{liu2023ingvio}
C.~Liu, C.~Jiang, and H.~Wang, ``Ingvio: A consistent invariant filter for fast and high-accuracy gnss-visual-inertial odometry,'' \emph{IEEE Robotics and Automation Letters}, vol.~8, no.~3, pp. 1850--1857, 2023.

\bibitem{niu2022ic}
X.~Niu, H.~Tang, T.~Zhang, J.~Fan, and J.~Liu, ``Ic-gvins: A robust, real-time, ins-centric gnss-visual-inertial navigation system,'' \emph{IEEE Robotics and Automation Letters}, vol.~8, no.~1, pp. 216--223, 2022.

\bibitem{cao2022gvins}
S.~Cao, X.~Lu, and S.~Shen, ``Gvins: Tightly coupled gnss--visual--inertial fusion for smooth and consistent state estimation,'' \emph{IEEE Transactions on Robotics}, vol.~38, no.~4, pp. 2004--2021, 2022.

\bibitem{li2022p}
T.~Li, L.~Pei, Y.~Xiang, W.~Yu, and T.-K. Truong, ``P$^3$-vins: Tightly-coupled ppp/ins/visual slam based on optimization approach,'' \emph{IEEE Robotics and Automation Letters}, vol.~7, no.~3, pp. 7021--7027, 2022.

\bibitem{xia2024invariant}
C.~Xia, X.~Li, S.~Li, and Y.~Zhou, ``Invariant-ekf-based gnss/ins/vision integration with high convergence and accuracy,'' \emph{IEEE/ASME Transactions on Mechatronics}, 2024.

\bibitem{chen2024end}
L.~Chen, P.~Wu, K.~Chitta, B.~Jaeger, A.~Geiger, and H.~Li, ``End-to-end autonomous driving: Challenges and frontiers,'' \emph{IEEE Transactions on Pattern Analysis and Machine Intelligence}, 2024.

\bibitem{mallik2023paving}
M.~Mallik, A.~K. Panja, and C.~Chowdhury, ``Paving the way with machine learning for seamless indoor--outdoor positioning: A survey,'' \emph{Information Fusion}, vol.~94, pp. 126--151, 2023.

\bibitem{xu2024machine}
P.~Xu, G.~Zhang, B.~Yang, and L.-T. Hsu, ``Machine learning in gnss multipath/nlos mitigation: Review and benchmark,'' \emph{IEEE Aerospace and Electronic Systems Magazine}, 2024.

\bibitem{siemuri2021machine}
A.~Siemuri, H.~Kuusniemi, M.~S. Elmusrati, P.~V{\"a}lisuo, and A.~Shamsuzzoha, ``Machine learning utilization in gnss—use cases, challenges and future applications,'' in \emph{2021 International Conference on Localization and GNSS (ICL-GNSS)}.\hskip 1em plus 0.5em minus 0.4em\relax IEEE, 2021, pp. 1--6.

\bibitem{sun2023resilient}
R.~Sun, L.~Fu, Q.~Cheng, K.-W. Chiang, and W.~Chen, ``Resilient pseudorange error prediction and correction for gnss positioning in urban areas,'' \emph{IEEE Internet of Things Journal}, vol.~10, no.~11, pp. 9979--9988, 2023.

\bibitem{zhang2024reliable}
X.~Zhang, X.~Wang, W.~Liu, X.~Tao, Y.~Gu, H.~Jia, and C.~Zhang, ``A reliable nlos error identification method based on lightgbm driven by multiple features of gnss signals,'' \emph{Satellite Navigation}, vol.~5, no.~1, pp. 1--20, 2024.

\bibitem{zhang2021prediction}
G.~Zhang, P.~Xu, H.~Xu, and L.-T. Hsu, ``Prediction on the urban gnss measurement uncertainty based on deep learning networks with long short-term memory,'' \emph{IEEE Sensors Journal}, vol.~21, no.~18, pp. 20\,563--20\,577, 2021.

\bibitem{zheng2024improving}
S.~Zheng, K.~Zeng, Z.~Li, Q.~Wang, K.~Xie, M.~Liu, and S.~Xie, ``Improving the prediction of gnss satellite visibility in urban canyons based on a graph transformer,'' \emph{NAVIGATION: Journal of the Institute of Navigation}, vol.~71, no.~4, 2024.

\bibitem{xu2024framework}
P.~Xu, G.~Zhang, Y.~Zhong, B.~Yang, and L.-T. Hsu, ``A framework for graphical gnss multipath and nlos mitigation,'' \emph{IEEE Transactions on Intelligent Transportation Systems}, 2024.

\bibitem{Zhang2022CTFGO}
H.~Zhang, X.~Xia, M.~Nitsch, and D.~Abel, ``Continuous-time factor graph optimization for trajectory smoothness of gnss/ins navigation in temporarily gnss-denied environments,'' \emph{IEEE Robotics and Automation Letters}, vol.~7, no.~4, pp. 9115--9122, 2022.

\bibitem{hu2024pyrtklib}
R.~Hu, P.~Xu, Y.~Zhong, and W.~Wen, ``pyrtklib: An open-source package for tightly coupled deep learning and gnss integration for positioning in urban canyons,'' \emph{arXiv preprint arXiv:2409.12996}, 2024.

\bibitem{mohanty2023learning}
A.~Mohanty and G.~Gao, ``Learning gnss positioning corrections for smartphones using graph convolution neural networks,'' \emph{NAVIGATION: Journal of the Institute of Navigation}, vol.~70, no.~4, 2023.

\bibitem{mohanty2024tightly}
------, ``Tightly coupled graph neural network and kalman filter for smartphone positioning,'' \emph{NAVIGATION: Journal of the Institute of Navigation}, vol.~71, no.~4, 2024.

\bibitem{ding2022learning}
Y.~Ding, P.~Chauchat, G.~Pages, and P.~Asseman, ``Learning-enhanced adaptive robust gnss navigation in challenging environments,'' \emph{IEEE Robotics and Automation Letters}, vol.~7, no.~4, pp. 9905--9912, 2022.

\bibitem{revach2022kalmannet}
G.~Revach, N.~Shlezinger, X.~Ni, A.~L. Escoriza, R.~J. Van~Sloun, and Y.~C. Eldar, ``Kalmannet: Neural network aided kalman filtering for partially known dynamics,'' \emph{IEEE Transactions on Signal Processing}, vol.~70, pp. 1532--1547, 2022.

\bibitem{niu2024mgins}
X.~Niu, L.~Ding, Y.~Wang, and J.~Kuang, ``Mgins: A lane-level localization system for challenging urban environments using magnetic field matching/gnss/ins fusion,'' \emph{IEEE Transactions on Intelligent Transportation Systems}, 2024.

\bibitem{bevis1994gps}
M.~Bevis, S.~Businger, S.~Chiswell, T.~A. Herring, R.~A. Anthes, C.~Rocken, and R.~H. Ware, ``Gps meteorology: Mapping zenith wet delays onto precipitable water,'' \emph{Journal of Applied Meteorology (1988-2005)}, pp. 379--386, 1994.

\bibitem{klobuchar1987ionospheric}
J.~A. Klobuchar, ``Ionospheric time-delay algorithm for single-frequency gps users,'' \emph{IEEE Transactions on aerospace and electronic systems}, no.~3, pp. 325--331, 1987.

\bibitem{zhang2024ins}
T.~Zhang, L.~Zhou, X.~Feng, J.~Shi, Q.~Zhang, and X.~Niu, ``Ins aided gnss pseudo-range error prediction using machine learning for urban vehicle navigation,'' \emph{IEEE Sensors Journal}, 2024.

\bibitem{zhang2023learning}
H.~Zhang, Z.~Wang, and H.~Vallery, ``Learning-based nlos detection and uncertainty prediction of gnss observations with transformer-enhanced lstm network,'' in \emph{2023 IEEE 26th International Conference on Intelligent Transportation Systems (ITSC)}.\hskip 1em plus 0.5em minus 0.4em\relax IEEE, 2023, pp. 910--917.

\bibitem{hu2023fisheye}
R.~Hu, W.~Wen, and L.-T. Hsu, ``Fisheye camera aided gnss nlos detection and learning-based pseudorange bias correction for intelligent vehicles in urban canyons,'' in \emph{2023 IEEE 26th International Conference on Intelligent Transportation Systems (ITSC)}.\hskip 1em plus 0.5em minus 0.4em\relax IEEE, 2023, pp. 6088--6095.

\bibitem{great2024dataset}
{GREAT (GNSS+ REsearch, Application and Teaching) Group from SGG of Wuhan University}, ``{GREAT Dataset: A vehicle-mounted multi-sensor raw observation dataset in complex urban environment},'' {Online}, 2024.

\bibitem{zhu2024large}
F.~Zhu, X.~Chen, Q.~Cai, and X.~Zhang, ``A large-scale diverse gnss/sins dataset: construction, publication, and application,'' \emph{IEEE Transactions on Instrumentation and Measurement}, 2024.

\end{thebibliography}

\begin{IEEEbiography}[{\includegraphics[width=1in,height=1.25in,clip,keepaspectratio]{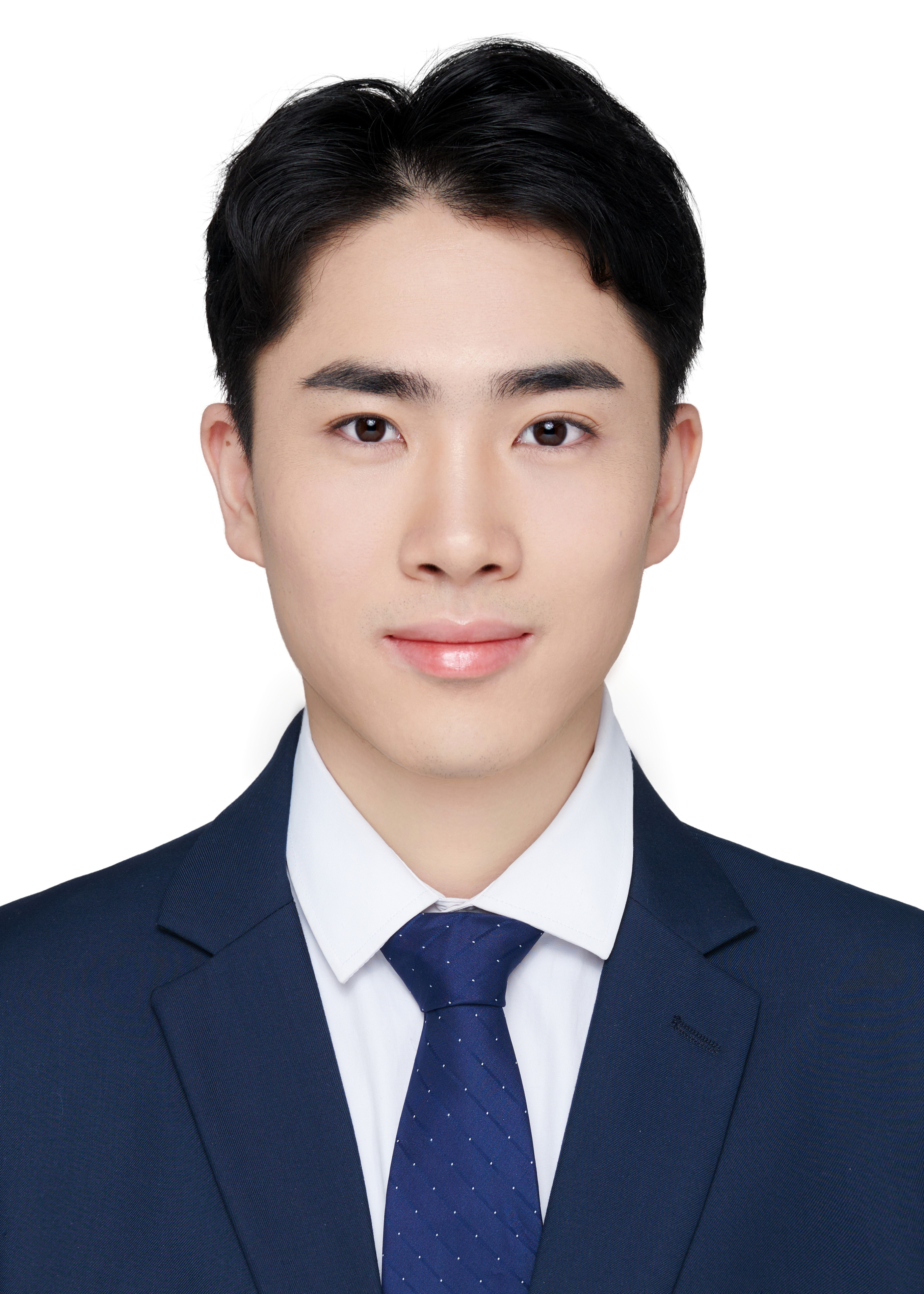}}]{Jianan Lou} received the B.Eng. degree in Navigation Engineering and the B.Ec. degree in Economics from Wuhan University, Wuhan, China, in 2022.

He is currently pursuing his Ph.D. degree at the Department of Precision Instrument and the State Key Laboratory of Precision Space-time Information Sensing Technology, Tsinghua University, Beijing, China. His research focuses on AI-enhanced multi-source fusion navigation and positioning technology. He has been an IEEE Student Member since 2023.
\end{IEEEbiography}

\begin{IEEEbiography}[{\includegraphics[width=1in,height=1.25in,clip,keepaspectratio]{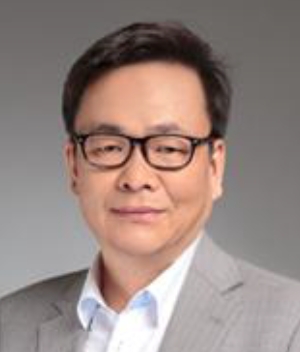}}]{Rong Zhang} received the B.S. degree in Mechanical Manufacturing and Automation from Tsinghua University, Beijing, China, in 1992. He received the M.S. degree in Precision Instrument and Machinery from Tsinghua University, and joined the faculty of Tsinghua University in 1994. In 2007, he received the Ph.D. degree in Precision Instrument and Machinery from Tsinghua University, and became a Professor in the Department of Precision Instrument, Tsinghua University. He is now the chief of the State Key Laboratory of Precision Space-time Information Sensing Technology, Tsinghua University. His professional interests include the precision motion control system, MEMS inertial sensors, and integrated navigation systems.

\end{IEEEbiography}
\end{document}